\documentclass[acmsmall, screen]{acmart}

\usepackage{multirow}
\usepackage{makecell}
\usepackage{graphicx}
\usepackage{pifont}
\usepackage{array}

\AtBeginDocument{%
  }

\setcopyright{acmlicensed}
\copyrightyear{2024}
\acmYear{2024}
\acmDOI{XXXXXXX.XXXXXXX}

\acmISBN{978-1-4503-XXXX-X/18/06}

\begin{document}

\title{Controllable Text Generation for Large Language Models: A Survey}

\author{Xun Liang}
\authornote{Both authors contributed equally to this research.}
\affiliation{%
  \institution{Renmin University of China}
  \city{Beijing}
  \country{China}
}
\author{Hanyu Wang}
\authornotemark[1]
\affiliation{%
  \institution{Renmin University of China}
  \city{Beijing}
  \country{China}
}
\author{Yezhaohui Wang}
\authornotemark[1]
\affiliation{%
  \institution{Institute for Advanced Algorithms Research, Shanghai}
  \city{Shanghai}
  \country{China}}
\author{Shichao Song}
\affiliation{%
  \institution{Renmin University of China}
  \city{Beijing}
  \country{China}
}
\author{Jiawei Yang}
\affiliation{%
  \institution{Renmin University of China}
  \city{Beijing}
  \country{China}
}

\author{Simin Niu}
\affiliation{%
  \institution{Renmin University of China}
  \city{Beijing}
  \country{China}
}
\author{Jie Hu}
\affiliation{%
  \institution{China Telecom Research Institute}
  \city{Beijing}
  \country{China}
}
\author{Dan Liu}
\affiliation{%
  \institution{China Telecom Research Institute}
  \city{Beijing}
  \country{China}
}
\author{Shunyu Yao}
\affiliation{%
  \institution{China Telecom Research Institute}
  \city{Beijing}
  \country{China}
}

\author{Feiyu Xiong}
\affiliation{%
  \institution{Institute for Advanced Algorithms Research, Shanghai}
  \city{Shanghai}
  \country{China}}
  
\author{Zhiyu Li}
\authornote{Corresponding author: lizy@iaar.ac.cn.}
\affiliation{%
  \institution{Institute for Advanced Algorithms Research, Shanghai}
  \city{Shanghai}
  \country{China}}

\renewcommand{\shortauthors}{Liang et al.}

\begin{abstract}

In Natural Language Processing (NLP), Large Language Models (LLMs) have demonstrated high text generation quality. However, in real-world applications, LLMs must meet increasingly complex requirements. Beyond avoiding misleading or inappropriate content, LLMs are also expected to cater to specific user needs, such as imitating particular writing styles or generating text with poetic richness. These varied demands have driven the development of Controllable Text Generation (CTG) techniques, which ensure that outputs adhere to predefined control conditions—such as safety, sentiment, thematic consistency, and linguistic style—while maintaining high standards of helpfulness, fluency, and diversity.

This paper systematically reviews the latest advancements in CTG for LLMs, offering a comprehensive definition of its core concepts and clarifying the requirements for control conditions and text quality. We categorize CTG tasks into two primary types: content control and attribute control. The key methods are discussed, including model retraining, fine-tuning, reinforcement learning, prompt engineering, latent space manipulation, and decoding-time intervention. We analyze each method’s characteristics, advantages, and limitations, providing nuanced insights for achieving generation control. Additionally, we review CTG evaluation methods, summarize its applications across domains, and address key challenges in current research, including reduced fluency and practicality. We also propose several appeals, such as placing greater emphasis on real-world applications in future research. This paper aims to offer valuable guidance to researchers and developers in the field. Our reference list and  Chinese version are open-sourced at \href{https://github.com/IAAR-Shanghai/CTGSurvey}{https://github.com/IAAR-Shanghai/CTGSurvey}.

\textcolor{red}{Note: This document, for the purpose of illustrating tasks related to safety in CTG, may contain examples that are offensive. Please read selectively.}

\end{abstract}



\received{XX XX XXXX}
\received[revised]{XX XX XXXX}
\received[accepted]{XX XX XXXX}

\maketitle

\section{Introduction}

With the rapid development of Large Language Models (LLMs) and their widespread application in Natural Language Processing (NLP), significant breakthroughs in text generation quality have been achieved \cite{zhao_arxiv23_LLMSurvey}. However, in practical applications, LLMs are often confronted with more complex and stringent content generation requirements. For example, in domains such as finance \cite{lee_arxiv24_surveyFinLLMs} and news reporting \cite{liang_arxiv24_UHGEval}, models must not only avoid generating misleading or discriminatory content \cite{bender_FACCT21_dangers}, but also precisely match specific conditions and user demands. These demands might include imitating a particular writing style or producing text with poetic qualities. Such requirements have driven the development of Controllable Text Generation (CTG) technologies, also known as Controlled Text Generation or Constrained Text Generation, which ensure that generated text meets both high-quality standards and the specific needs of various applications.

\begin{figure}[h!]
    \centering
    \includegraphics[width=0.65\linewidth]{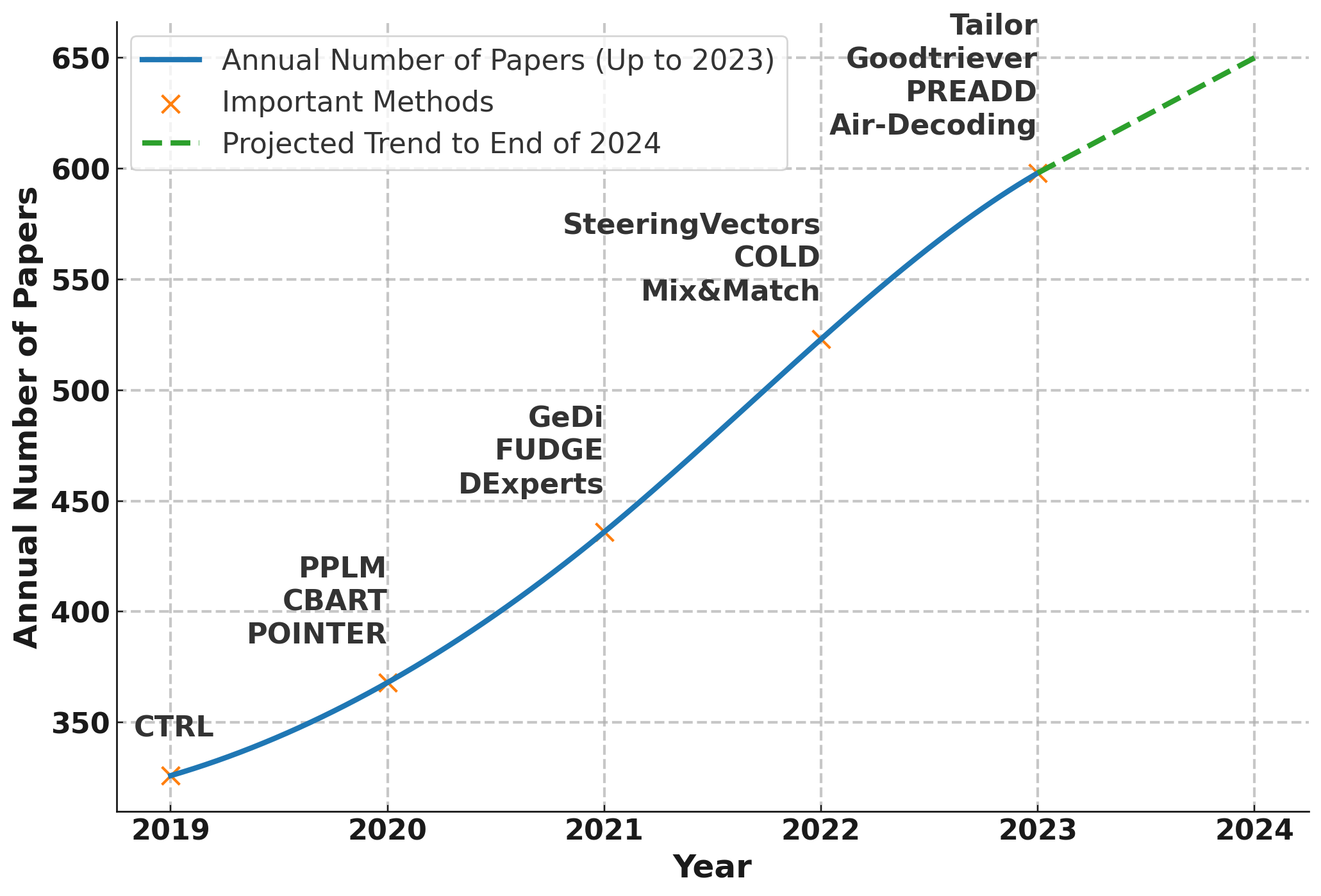}
    \caption{Publication trends on Web of Science related to Controllable Generation in Language Models.}
    \label{fig:trends}
\end{figure}

The increasing interest and demand for enabling LLMs to generate content that meets specific requirements have driven the expansion of CTG research. Figure~\ref{fig:trends} illustrates the growth in the number of papers related to "Control Generation in Language Models" indexed by Web of Science\footnote{\url{https://www.webofscience.com}}.

CTG guides text generation to follow predefined control conditions, such as safety or sentiment, while maintaining quality like fluency and diversity \cite{zhang_ACMCS23_CTGSurvey}. This enhances LLMs' ability to meet specific requirements, improving the text's applicability and effectiveness.

Control conditions in CTG can be explicit or implicit. Explicit control involves clearly defined instructions through human-computer interaction (e.g., input prompts), directing the model to generate text in a specific style, such as in a Shakespearean or humorous tone \cite{tao_arxiv24_CAT}. Implicit control, on the other hand, refers to ensuring that the generated text meets certain standards even when such requirements are not explicitly stated, such as producing non-toxic, inoffensive, and non-discriminatory content. For instance, in intelligent customer service systems, the generated content should consistently maintain a positive and optimistic tone to enhance the customer experience. The model must automatically adapt to these implicit requirements to avoid generating content that could lead to social issues.

\begin{figure}[h]
    \centering
    \includegraphics[width=0.8\textwidth]{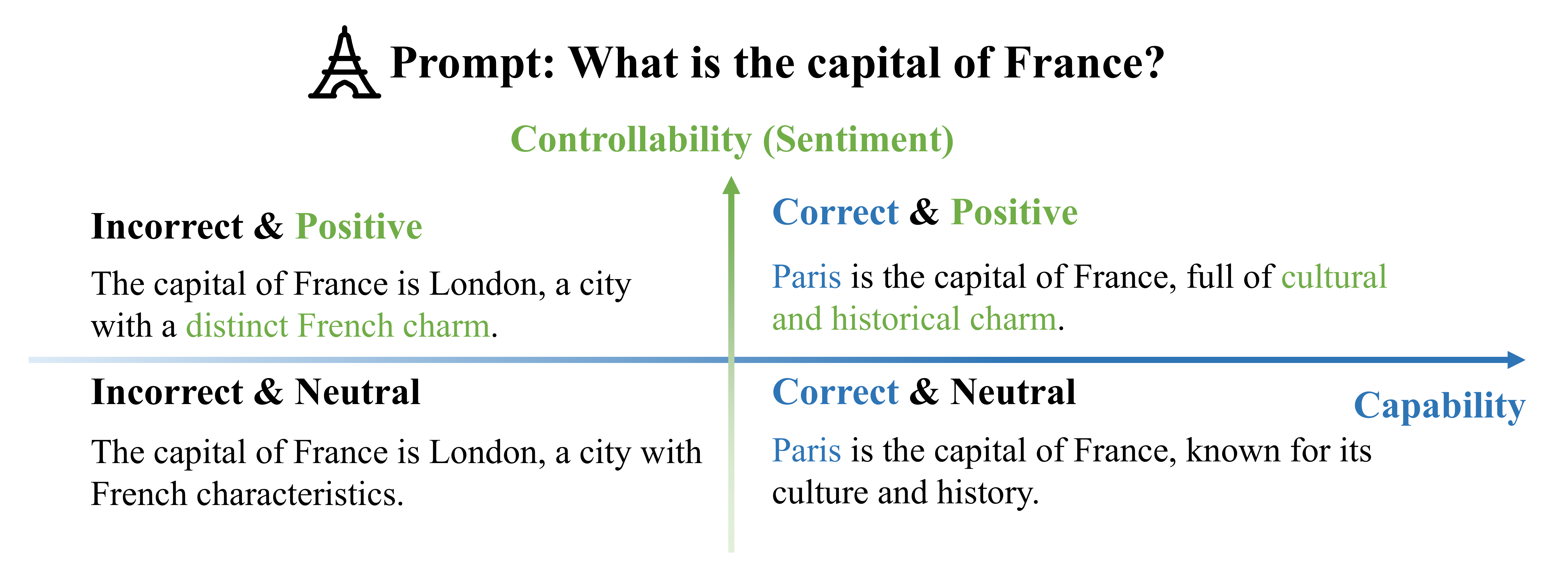}
    \caption{Controllability dimension and capability dimension of LLMs.}
    \label{fig:controllability_vs_capability}
\end{figure}

CTG can be considered an ability dimension orthogonal to the objective knowledge capabilities of LLMs. As illustrated in Figure \ref{fig:controllability_vs_capability}, while LLMs excel in objective capabilities such as logical reasoning, text analysis, or problem-solving \cite{liang_arxiv24_ICSF}, CTG emphasizes the manner in which this objective information is expressed and presented. In other words, CTG not only focuses on the accuracy and relevance of the facts in the generated text but also places special importance on how this information is conveyed.
For example, in sentiment control, CTG does not require the model to prioritize the factual accuracy of the content but instead ensures that the sentiment conveyed aligns with the intended emotional tone. Similarly, in style control, the model must ensure that the content adheres to a specific linguistic style or tone. CTG empowers LLMs to generate more personalized and context-sensitive content that meets varying user requirements.
It is important to recognize, however, that there is no absolute standard dictating that positive sentiment output is inherently superior to neutral sentiment output. The focus of CTG tasks lies in adapting to different application scenarios and requirements to achieve the most suitable generation outcome.

\subsection{Demands of Controllable Text Generation}
\label{sec:ctg_dim}

The demands of CTG can be categorized into two primary dimensions. The first involves ensuring that the generated text conforms to predefined control conditions, such as text structure, safety, and thematic focus, to meet user needs. The second dimension focuses on maintaining the text's helpfulness, fluency, and diversity as fundamental quality standards, ensuring its effectiveness and applicability in real-world scenarios. Together, these dimensions present a dual challenge in CTG: rigorously adhering to specified control conditions while upholding high standards of text quality.

\subsubsection{\textbf{Dimension 1: Meeting Predefined Control Conditions}}
The primary objective of CTG is to ensure that the generated text adheres to predefined control conditions. This involves tailoring the text to meet specific objectives or requirements, making it well-suited for its intended application. Control conditions may include generating text on a particular topic, ensuring safety by avoiding harmful content, or emulating specific linguistic styles.

For example, in terms of safety, the model must avoid generating content that could be perceived as harmful, such as discriminatory or violent language. Consider the following scenario:
\begin{itemize}
    \item \textit{Original Input:} "His child is really stupid."
    \item \textit{Controlled Output:} "It's wrong to say that; it could cause harm."
\end{itemize}

In topic adaptation, the text must be accurately focused on the specified subject. For example:
\begin{itemize}
    \item \textit{Original Input:} "With the rapid changes in the economy, today's market has shown unprecedented dynamics. Investors and consumers are monitoring trends to make decisions."
    \item \textit{Finance-themed:} "In this context, the stock market quickly reacted after the morning bell, with major indices like the Dow Jones Industrial Average and NASDAQ seeing slight increases."
    \item \textit{Sports-themed:} "In such an economic environment, the sports industry has also demonstrated unique adaptability. Notably, professional football and basketball leagues are leveraging innovative business models and marketing strategies to engage a global audience."
\end{itemize}

Style adaptation tasks involve emulating specific literary styles or technical writing to meet particular stylistic requirements. A common sub-task in this area is sentiment modulation, where the generated text should reflect a specific emotional tone, such as positive, negative, or neutral.

A more detailed discussion of these controllable tasks and their applications can be found in Section \ref{sec:ctg_tasks}, which elaborates on specific scenarios and implementations.

\subsubsection{\textbf{Dimension 2: Maintaining Text Quality}}
In addition to achieving predefined control conditions, it is essential to maintain its fluency, helpfulness, and diversity.

\begin{itemize}
    \item \textbf{Fluency:} The text should be smooth and logically coherent, making it easy for the reader to understand the intent and context. Consider the following example:
    \begin{itemize}
        \item \textit{Original Input:} "I need advice on growing tomatoes."
        \item \textit{Output 1:} "Tomatoes need sun water good grow." (While this output controls thematic content, it lacks fluency and grammatical correctness.)
        \item \textit{Output 2:} "Tomatoes grow well in sunny locations and require regular watering." (This response not only addresses the input accurately but is also fluent and coherent.)
    \end{itemize}
    \item \textbf{Helpfulness:} The text generated by LLMs should provide real-world value, helping to solve specific problems or offering necessary information. Consider the following example:
    \begin{itemize}
        \item \textit{Original Input:} "His child is really stupid."
        \item \textit{Output 1:} "I can't answer." (This response avoids harm but lacks constructive advice.)
        \item \textit{Output 2:} "You shouldn't say that. Everyone has areas they can improve. Let's identify the issue and find a solution." (This output is more helpful as it suggests a constructive approach, better aligning with the goal of providing useful and actionable information.)
    \end{itemize}
    \item \textbf{Diversity:} The text should avoid being repetitive or formulaic. Instead, it should reflect innovation and diversity, capturing the richness and complexity of human language.
\end{itemize}

\begin{figure}[htbp]
    \centering
    \includegraphics[width=\textwidth]{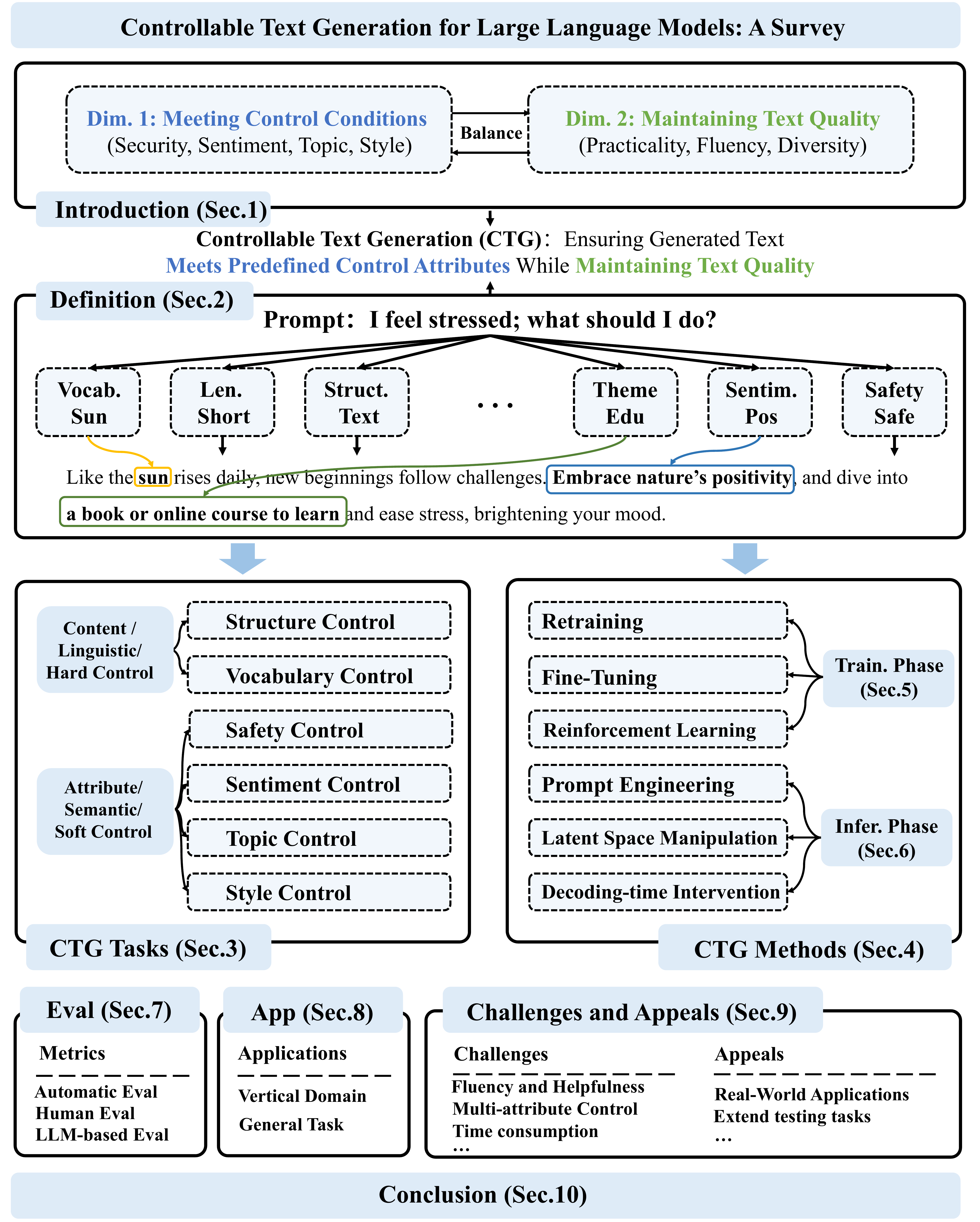}
    \caption{Survey Framework}
    \label{fig:framework}
\end{figure}

\subsection{Related Surveys}

CTG has been extensively explored in recent years. Table \ref{tab:survey} summarizes key surveys in CTG.

\textit{Exploring Controllable Text Generation Techniques} \cite{prabhumoye_COLING20_CTGSurvey} is one of the earliest surveys in this area, providing a general framework that covers techniques across various model architectures, including RNNs \cite{rumelhart_1986_rnn}, LSTMs \cite{hochreiter_1997_lstm}, and Transformers \cite{vaswani_nips17_transformer}.

\textit{Conditional Text Generation for Harmonious Human-Machine Interaction} \cite{guo_TIST21_HMI} examines CTG from a practical application perspective, particularly in human-machine interaction. This survey emphasizes sentiment and personalized text generation, using models like RNNs \cite{rumelhart_1986_rnn}, LSTMs \cite{hochreiter_1997_lstm}, GANs \cite{DBLP:journals/corr/RadfordMC15}, Transformers \cite{vaswani_nips17_transformer}, and VAEs \cite{kingma_arxiv13_vae}, with a strong focus on real-world applications.

\textit{How to Control Sentiment in Text Generation: A Survey of the State-of-the-Art in Sentiment-Control Techniques} \cite{lorandi_WASSA23_sentimentCTG} provides an in-depth look at sentiment control within CTG, highlighting the challenges and importance of managing sentiment in generated text.

\textit{A Recent Survey on Controllable Text Generation: A Causal Perspective} \cite{wang_2024_CausalCTG} critiques traditional CTG methods focused on statistical correlations, advocating for improvements via representation disentanglement, causal inference, and knowledge augmentation.

\textit{A Survey of Controllable Text Generation using Transformer-based Pre-trained Language Models} \cite{zhang_ACMCS23_CTGSurvey} focuses on Transformer-based pre-trained models in CTG. While it discusses the evolving capabilities and limitations of these models, it also addresses challenges in systematically categorizing CTG tasks and methods. For example, tasks like table-to-text generation may blur the lines between general language modeling and CTG-specific tasks. Additionally, the classification of prompts under fine-tuning methods suggests a need for clearer distinctions as CTG methodologies evolve. Due to the rapid advancements in LLMs and emerging methods like latent space manipulation in 2023 and 2024, the survey's pre-2022 references may be less relevant for current LLM research.

\begin{table}[htbp]
\centering
\caption{Summary of Surveys in Controllable Text Generation}
\label{tab:survey}
\footnotesize
\begin{tabular}{llllllll}
\hline
\textbf{Surveys} &  & \textbf{\cite{prabhumoye_COLING20_CTGSurvey}} & \textbf{\cite{guo_TIST21_HMI}} & \textbf{\cite{lorandi_WASSA23_sentimentCTG}} & \textbf{\cite{wang_2024_CausalCTG}} & \textbf{\cite{zhang_ACMCS23_CTGSurvey}} & \textbf{Ours} \\ \hline
\multirow{2}{*}{\textbf{Models}} & PLMs & \multicolumn{1}{c}{\ding{51}} & \multicolumn{1}{c}{\ding{51}} & \multicolumn{1}{c}{\ding{51}} & \multicolumn{1}{c}{\ding{51}} & \multicolumn{1}{c}{\ding{51}} & \multicolumn{1}{c}{\ding{51}} \\
& LLMs (Large-scale PLMs \cite{zhao_arxiv23_LLMSurvey}) &  &  &  & \multicolumn{1}{c}{\ding{51}} &  & \multicolumn{1}{c}{\ding{51}} \\ \hline
\multirow{2}{*}{\textbf{Tasks}} & Abstract Attributes & \multicolumn{1}{c}{\ding{51}} & \multicolumn{1}{c}{\ding{51}} & \multicolumn{1}{c}{\ding{51}} & \multicolumn{1}{c}{\ding{51}} & \multicolumn{1}{c}{\ding{51}} & \multicolumn{1}{c}{\ding{51}} \\
& Concrete Attributes &  &  &  &  & \multicolumn{1}{c}{\ding{51}} & \multicolumn{1}{c}{\ding{51}} \\ \hline
\multirow{3}{*}{\textbf{\makecell[l]{Learning-Based Methods}}} & Training & \multicolumn{1}{c}{\ding{51}} & \multicolumn{1}{c}{\ding{51}} & \multicolumn{1}{c}{\ding{51}} & \multicolumn{1}{c}{\ding{51}} & \multicolumn{1}{c}{\ding{51}} & \multicolumn{1}{c}{\ding{51}} \\
& Fine-Tuning &  &  & \multicolumn{1}{c}{\ding{51}} & \multicolumn{1}{c}{\ding{51}} & \multicolumn{1}{c}{\ding{51}} & \multicolumn{1}{c}{\ding{51}} \\
& Reinforcement Learning &  &  &  &  & \multicolumn{1}{c}{\ding{51}} & \multicolumn{1}{c}{\ding{51}} \\ \hline
\multirow{3}{*}{\textbf{\makecell[l]{Unlearning Methods}}} & Input Optimization & \multicolumn{1}{c}{\ding{51}} &  &  & \multicolumn{1}{c}{\ding{51}} & \multicolumn{1}{c}{\ding{51}} & \multicolumn{1}{c}{\ding{51}} \\
&  Internal Processing Manipulation &  &  &  &  &  & \multicolumn{1}{c}{\ding{51}} \\
& Output Intervention & \multicolumn{1}{c}{\ding{51}} & \multicolumn{1}{c}{\ding{51}} & \multicolumn{1}{c}{\ding{51}} &  & \multicolumn{1}{c}{\ding{51}} & \multicolumn{1}{c}{\ding{51}} \\ \hline
\multirow{3}{*}{\textbf{\makecell[l]{Evaluation Methods}}} & General Metrics &  & \multicolumn{1}{c}{\ding{51}} & \multicolumn{1}{c}{\ding{51}} & \multicolumn{1}{c}{\ding{51}} & \multicolumn{1}{c}{\ding{51}} & \multicolumn{1}{c}{\ding{51}} \\
& Task-specific Metrics &  & \multicolumn{1}{c}{\ding{51}} & \multicolumn{1}{c}{\ding{51}} & \multicolumn{1}{c}{\ding{51}} & \multicolumn{1}{c}{\ding{51}} & \multicolumn{1}{c}{\ding{51}} \\
& Benchmarks &  &  &  &  &  & \multicolumn{1}{c}{\ding{51}} \\ \hline
\multirow{2}{*}{\textbf{Applications}} & Horizontal Applications &  & \multicolumn{1}{c}{\ding{51}} &  &  & \multicolumn{1}{c}{\ding{51}} & \multicolumn{1}{c}{\ding{51}} \\
& Vertical Applications &  &  &  &  &  & \multicolumn{1}{c}{\ding{51}} \\ \hline
\multirow{4}{*}{\textbf{\makecell[l]{Discussions}}} & Control Mechanisms in CTG & \multicolumn{1}{c}{\ding{51}}  &  &  &  &  & \multicolumn{1}{c}{\ding{51}} \\
& Quality of Control in CTG &  &  &  & \multicolumn{1}{c}{\ding{51}}  &  & \multicolumn{1}{c}{\ding{51}} \\
& Challenges in Current Methods & \multicolumn{1}{c}{\ding{51}} & \multicolumn{1}{c}{\ding{51}} & \multicolumn{1}{c}{\ding{51}} & \multicolumn{1}{c}{\ding{51}} & \multicolumn{1}{c}{\ding{51}} & \multicolumn{1}{c}{\ding{51}} \\
& Future Research Directions &  & \multicolumn{1}{c}{\ding{51}} &  & \multicolumn{1}{c}{\ding{51}} & \multicolumn{1}{c}{\ding{51}} & \multicolumn{1}{c}{\ding{51}} \\ \hline
\textbf{Cutoff Year for References} &  & 2020 & 2020 & 2022 & 2023 & 2022 & \textbf{2024} \\ \hline
\end{tabular}
\end{table}

The dimensions outlined in Table \ref{tab:survey} provide a comprehensive overview of key CTG surveys. These dimensions—ranging from model choice (from small-scale PLMs to large-scale LLMs as defined in \cite{zhao_arxiv23_LLMSurvey}), task categorization (abstract and concrete attribute control), learning methods (training, fine-tuning, reinforcement learning), unlearning methods (input optimization, internal manipulation, output intervention), evaluation criteria (general and task-specific metrics), to application scenarios (horizontal and vertical applications)—significantly influence the scope and depth of CTG research. Furthermore, discussions on control mechanisms, quality considerations, challenges, and future directions highlight the underlying mechanisms and potential of CTG. The inclusion of a reference cutoff year ensures that the latest developments are covered.

Compared to existing surveys, the core contributions and unique features of this review include:

\begin{itemize}
    \item \textbf{Focus on Transformer Architecture:} This paper explores the application of pre-trained LLMs based on the Transformer architecture \cite{vaswani_nips17_transformer} in CTG. While models like RNNs \cite{rumelhart_1986_rnn}, LSTMs \cite{hochreiter_1997_lstm}, and VAEs \cite{kingma_arxiv13_vae} have significantly contributed to CTG, our primary focus is on Transformer-based models, highlighting their advantages and applications in this field.

    \item \textbf{Emphasis on Large Language Models:} This paper centers on the latest advancements in CTG methods, particularly with the rise of large pre-trained language models such as GPT \cite{tom_nips20_gpt} and Llama \cite{touvron_arxiv23_llama}. The development and application of these LLMs in 2023 and 2024 have driven a wave of innovation in CTG, reshaping research perspectives. Consequently, this paper focuses on CTG methods tailored for large pre-trained language models in the LLM era, introducing the concepts and characteristics of these cutting-edge approaches.

    \item \textbf{Exploration of Model Expression and CTG Quality:} This paper examines the interplay between CTG and model capabilities, exploring how external control conditions are integrated into the CTG process. It also addresses the quality of CTG, focusing on what defines more effective and useful text generation. 

    \item \textbf{Innovative Task Classification Framework:} This paper introduces a novel framework for classifying CTG tasks into two primary categories: content control (hard control) and attribute control (soft control). This framework provides a structured approach to exploring and analyzing the diversity of CTG methods.
    
    \item \textbf{Systematic Classification of CTG Methods:} This paper categorizes CTG methods into two main stages: training-stage methods and inference-stage methods. These encompass techniques such as retraining, fine-tuning, reinforcement learning, prompt engineering, latent space manipulation, and decoding-time intervention.
\end{itemize}

\subsection{Paper Structure}

The logical framework of this paper is outlined in Figure \ref{fig:framework}. Section \ref{sec:ctg_dim} begins by introducing the core requirements of CTG. In Section \ref{sec:ctg_def}, we define CTG within the context of LLMs, explaining key concepts and exploring how control conditions are integrated into the generation process.

Section \ref{sec:ctg_tasks} categorizes CTG tasks into content control (or linguistic control/hard control) and attribute control (or semantic control/soft control).

To provide a comprehensive overview of CTG methods, Section \ref{sec:ctg_methods} systematically categorizes techniques, ranging from retraining and fine-tuning during the training phase to prompt engineering and latent space manipulation during inference. These are discussed in detail in Sections \ref{sec:train_methods} and \ref{sec:infer_methods}.

Section \ref{sec:ctg_eval} delves into evaluation standards, presenting prevalent evaluation frameworks and techniques. Section \ref{sec:ctg_app} explores practical applications of CTG across various domains, such as news generation, dialogue systems, and toxicity reduction.

In Section \ref{sec:ctg_c&a}, we discuss challenges in CTG, including precise content control, the complexity of multi-attribute control, and the enhancement of text fluency and helpfulness. We advocate for diversifying test tasks, emphasizing practical applications, and maximizing the capabilities of LLMs.

Finally, Section \ref{sec:conclusion} summarizes the key contributions of this research, offering valuable insights for future developments in the CTG field.

\section{Definition}
\label{sec:ctg_def}

\subsection{Fundamental Principles of Text Generation}
LLMs based on the Transformer architecture \cite{vaswani_nips17_transformer} generate text by computing the conditional probability of sequence elements. Specifically, these models generate text by determining the probability of each token given the preceding tokens. This process can be expressed as:

\begin{equation}
P(X) = P(x_1, x_2, \ldots, x_n) = \prod_{i=1}^{n} p(x_i | x_{<i})
\end{equation}

Here, \(x_i\) represents the token currently being generated, and \(x_{<i}\) includes all the preceding tokens in the sequence. This probabilistic framework enables LLMs to generate diverse, coherent, and contextually relevant text, ensuring that each new token logically aligns with the context established by the preceding sequence.

\subsection{Definition of Controllable Text Generation}
In CTG, the primary objective is to integrate control conditions \(C\) into the text generation process while preserving the original text quality \cite{zhang_ACMCS23_CTGSurvey}. These control conditions guide the model to generate text with specific attributes, such as emotional tone or toxicity level, to meet particular application needs. Simultaneously, it is essential to ensure that the generated text maintains high standards in quality dimensions such as fluency, coherence, and diversity. The mathematical expression for the controlled generation process is as follows:

\begin{equation}
P(X | C) = P(x_1, x_2, \ldots, x_n | C) = \prod_{i=1}^{n} p(x_i | x_{<i}, C)
\end{equation}

In this equation, \(C\) represents a set of desired attributes that the generated text should reflect. The primary challenge of CTG lies in seamlessly incorporating these control conditions \(C\) into the generation process without compromising the inherent quality of the LLMs' output.

\subsection{Semantic Space Representation of Controllable Text Generation}
The problem of CTG can be framed within an ideal semantic space \(\mathcal{S} \subset \mathbb{R}^d\) \cite{liang_arxiv24_DATG}, where the output of LLMs is represented as vectors in this semantic space. The ideal semantic space \(\mathcal{S}\) is a multidimensional vector space in which the language model operates to generate text, encompassing all possible semantic representations. This space \(\mathcal{S}\) is a subset of \(\mathbb{R}^d\), containing all potential semantic vectors that the model could generate.

In this semantic space, the attributes of generated text—such as sentiment, safety, fluency, and lexical constraints—can be effectively decoupled into distinct dimensions. The primary goal in CTG is to adjust specific dimensions related to control conditions \(C\) within this space, guiding the generated text toward desired attributes while preserving the integrity of other semantic aspects.

In CTG, these semantic vectors can be manipulated through a transformation function \(f\), which strategically adjusts the vectors to align with desired attributes without compromising other semantic qualities. The effectiveness of transformation is evaluated through an optimization objective, ensuring that the text attributes meet expectations while maintaining overall semantic integrity.

\begin{equation}
J(f) = \mathbb{E}_{\mathbf{x} \sim P(\mathcal{S})} [-s(f(\mathbf{x}))]
\end{equation}

Here, \(\mathbf{x}\) represents a semantic vector drawn from the distribution \(P(\mathcal{S})\), where \(P(\mathcal{S})\) denotes the probability distribution of vectors within the semantic space \(\mathcal{S}\). The function \(s(\cdot)\) is a scoring function used to evaluate how well the transformed vector \(f(\mathbf{x})\) aligns with the control conditions \(C\). The transformation function \(f\) is defined as:

\begin{equation}
\mathbf{x}_{\text{after}} = f(\mathbf{x}_{\text{before}}) = \mathbf{x}_{\text{before}} + \Delta\mathbf{x}
\end{equation}

In this equation, \(\mathbf{x}_{\text{before}}\) represents the original semantic vector, and \(\Delta\mathbf{x}\) is the adjustment applied to modify the text's semantic characteristics according to the attributes specified by \(C\). This adjustment reshapes the text distribution within the semantic space, ensuring that the fundamental properties of the original vector are preserved while aligning it with the desired attributes.

\section{Tasks in Controllable Text Generation}
\label{sec:ctg_tasks}

In the realm of CTG, tasks can be broadly categorized into two main types based on the nature of the text control: content control (or linguistic control/hard control) and attribute control (or semantic control/soft control).


\subsection{Content Control (or Linguistic Control/Hard Control)}
Content control (linguistic control or hard control) focuses on specific elements of the generated text, such as its structure and vocabulary. This type of control requires the model to generate text content precisely according to predefined rules, earning the term "hard control" because it directly influences the specific form and content of the generated text. This category includes:

\begin{itemize}
    \item \textbf{Structure Control:}
    \begin{itemize}
        \item \textit{Specific Formats:} Generating text that adheres to specific formatting requirements, such as poetry \cite{yang_acl21_fudge,zou_KDD21_Inverse-Prompting}, recipes \cite{liu_acl22_RecipeWithPlans}, or other types of structured text, each with its own unique language and structural norms.
        \item \textit{Organizational Structure:} Ensuring that the text has appropriate paragraph divisions, the use of headings, and list arrangements \cite{hua_emnlp20_PAIR,lin_nuse21_PlugandBlend} to enhance clarity and readability.
        \item \textit{Length Control:} Managing the overall length of the generated text to meet specific requirements \cite{chai_arxiv22_fast,juseon_acl24_instructcmp,jie_acl24_LengthPrompt}, ensuring its suitability for the intended platform or purpose.
    \end{itemize}
    \item \textbf{Vocabulary Control:}
    \begin{itemize}
        \item \textit{Keyword Inclusion:} Ensuring that the generated text includes a predefined set of keywords \cite{zhang_emnlp20_POINTER,he_emnlp21_CBART}, thereby meeting specific informational needs and enhancing the relevance and specificity of the presented information.
        \item \textit{Prohibition of Specific Terms:} Preventing the use of potentially harmful or inappropriate terms \cite{lu_acl22-neurologic-AFesque}, thus maintaining the integrity and appropriateness of the content.
    \end{itemize}
\end{itemize}

\subsection{Attribute Control (or Semantic Control/Soft Control)}
Attribute control, also known as semantic control or soft control, focuses on abstract language attributes of the text, such as sentiment, style, and topic. The goal of this type of control is to ensure that the generated text reflects specific semantic characteristics at a higher level, rather than strictly defining precise linguistic expressions. This type of control is termed "soft control" because it emphasizes influencing the overall abstract characteristics of the text rather than its specific content. Examples include:

\begin{itemize}
    \item \textbf{Safety Control:}
    \begin{itemize}
        \item \textit{Detoxification:} The generated text should avoid any form of harmful content \cite{liu_acl21_DExperts,schick_tacl21_SD,dai_iclr24_SafeRLHF}, such as discriminatory language or violent content.
        \item \textit{Compliance with Laws and Regulations:} The text must adhere to all applicable legal and regulatory requirements \cite{bai_arxiv22_constitutionalai}, including privacy protection and copyright laws.
    \end{itemize}
    
    \item \textbf{Sentiment Control:}
    \begin{itemize}
        \item \textit{Sentiment Orientation:} Ensuring that the generated text exhibits a clear sentiment orientation, such as positive, negative, or neutral, to match specific communication purposes \cite{dathathri_iclr20_PPLM,zeldes_arxiv20_AuxiliaryTuning,chan_iclr21_CoCon,krause_emnlp21_gedi}. This ensures that the emotional tone aligns with the context or intended impact on the audience.
    \end{itemize}
    
    \item \textbf{Style Control:}
    \begin{itemize}
        \item \textit{General Style:} General style control ensures that the generated text meets the needs of specific occasions and industries \cite{keskar_arxiv19_Ctrl}. For instance, in fields like medicine, law, or business, it is necessary to maintain professional communication styles to ensure content professionalism and adaptability. Additionally, in different social settings, the text should reflect specific tones, such as formality or politeness \cite{saha_ijcai22_CounterGeDi,trotta_gem22_kNN-SCG}, to meet etiquette requirements.
        
        \item \textit{Personal Style:} Personal style control involves generating text that mimics a specific writing style \cite{upadhyay_arxiv22_DRL,tao_arxiv24_CAT,subramani_acl22_LatentStreeringVectors}, such as the Shakespearean style, to meet artistic or professional demands. It also includes generating personalized text according to individual expression habits and preferences, providing a more customized user experience.
    \end{itemize}
    
    \item \textbf{Topic Control:}
    \begin{itemize}
        \item \textit{Thematic Consistency:} Ensuring that the text strictly adheres to the specified theme \cite{dathathri_iclr20_PPLM,chan_iclr21_CoCon}, such as technology, sports, or politics. This includes aligning the content with the expected knowledge and interests of the target audience.
    \end{itemize}
    
\end{itemize}

These examples represent common tasks and application scenarios in CTG. Within the domains of content control and attribute control, numerous other rich tasks exist, all contributing to the broader research area of CTG.

\section{Classification of Controllable Text Generation Methods}
\label{sec:ctg_methods}

\begin{figure}[ht]
    \centering
    \includegraphics[width=0.9\textwidth]{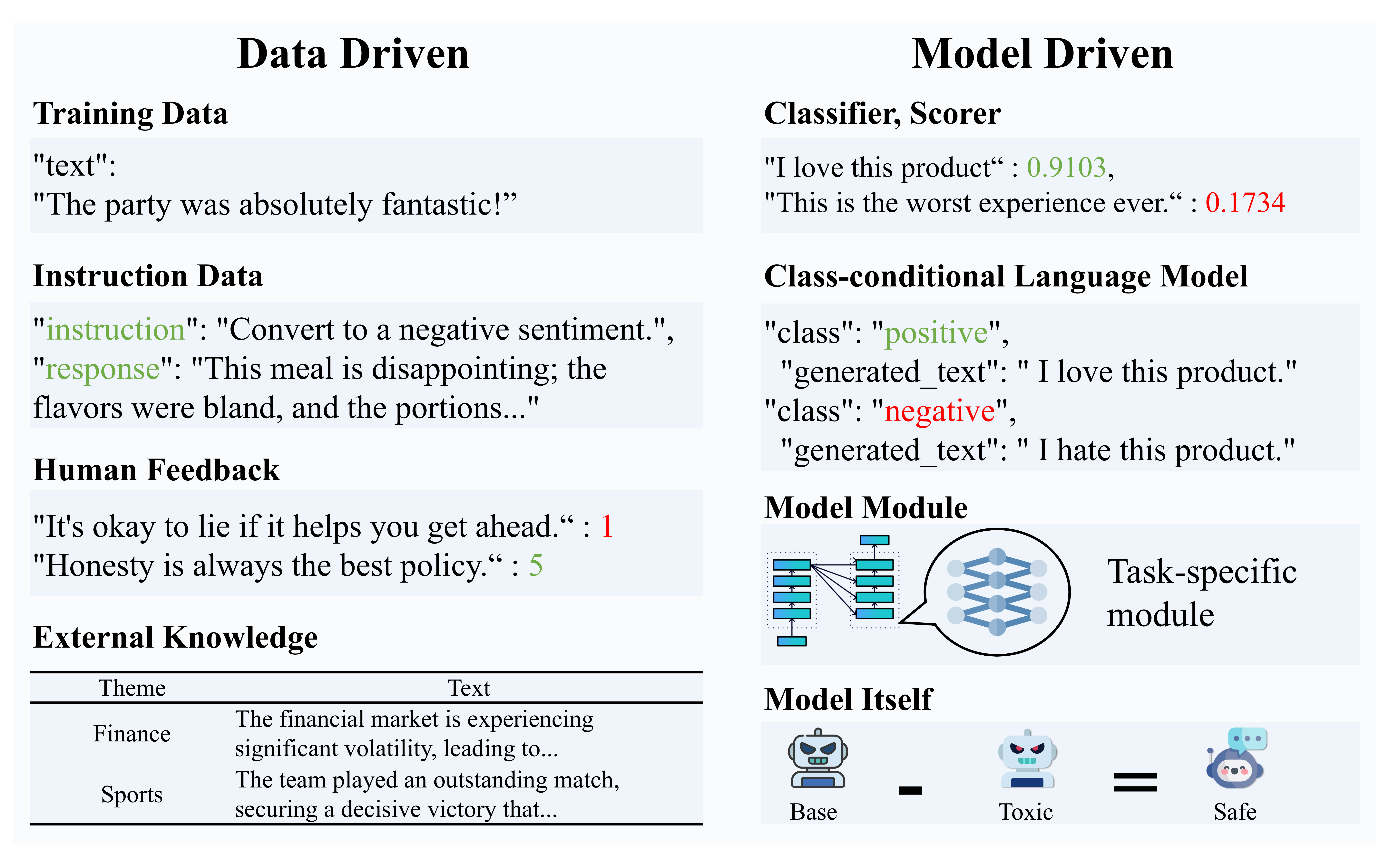}
    \caption{Injection of Conditions in CTG}
    \label{fig:injection}
\end{figure}

The core of CTG lies in integrating control conditions \(C\) into the text generation process of LLMs. CTG methods achieve this by injecting external information into the text generated by LLMs, either through parameterized or non-parameterized approaches. This external information can take various forms, including model-driven methods that utilize classifiers, conditional language models, or knowledge injection directly from the LLMs themselves. Alternatively, data-driven methods leverage rich data resources, such as text corpora \cite{keskar_arxiv19_Ctrl,zeldes_arxiv20_AuxiliaryTuning}, lexicons \cite{pascual_emnlp21_K2T}, graphs \cite{liang_arxiv24_DATG}, and databases \cite{nawezi_tllm23_kNN-CTG,pozzobon_emnlp23_goodtriever} to inject knowledge, as illustrated in Figure \ref{fig:injection}.The exact methodology and more details will be presented and discussed in Sections \ref{sec:train_methods} and \ref{sec:infer_methods}.

\begin{figure}[ht]
    \centering
    \includegraphics[width=0.8\textwidth]{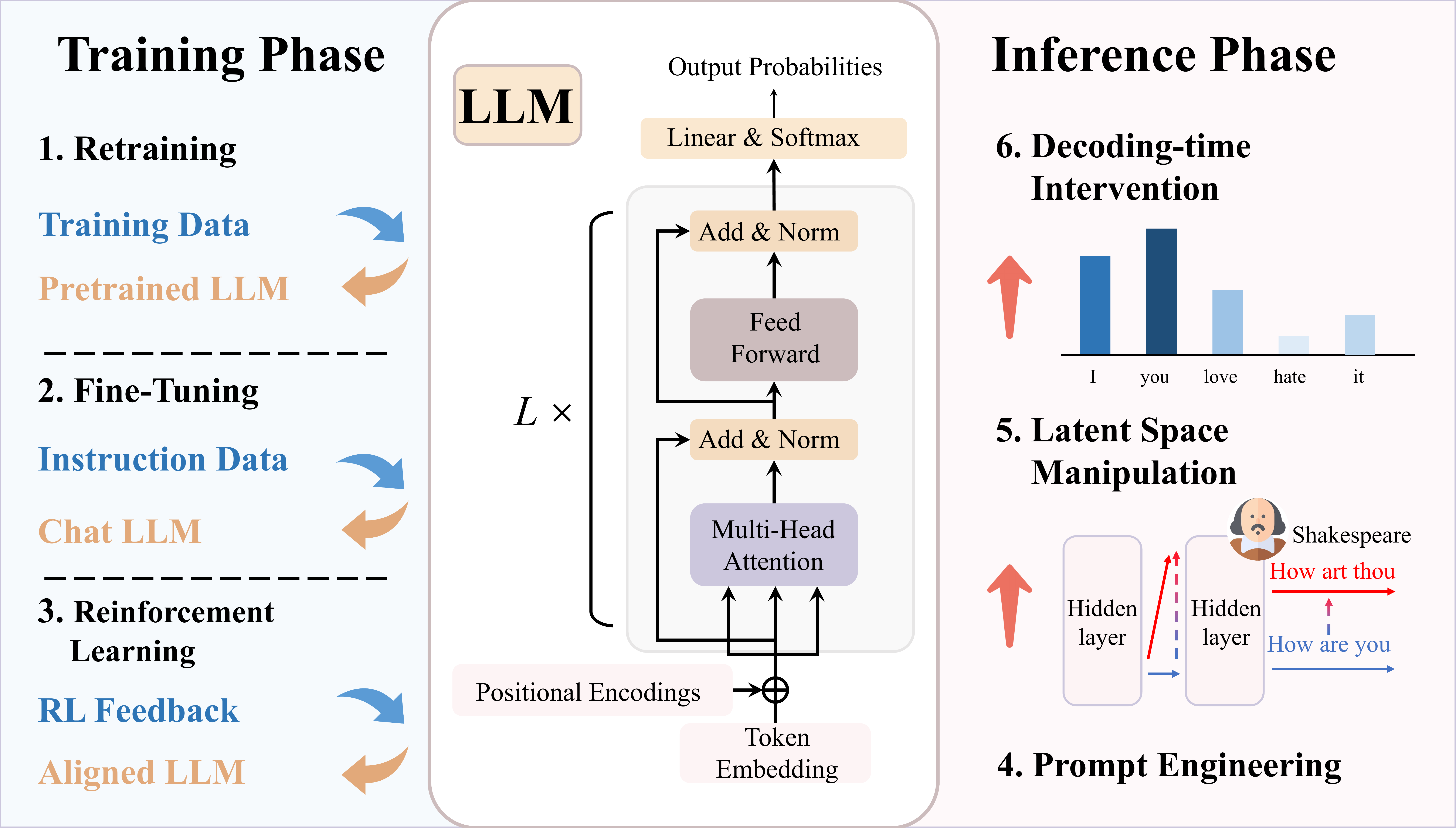}
    \caption{Classification of Controllable Text Generation Methods}
    \label{fig:methods_classification}
\end{figure}

CTG methods can be classified based on the stage at which model intervention occurs. Broadly, CTG methods are divided into two main stages: the training stage and the inference stage (see Figure \ref{fig:methods_classification}). Within each stage, CTG methods are further subdivided into different categories, as shown in Table \ref{tab:classification}, encompassing various research approaches and specific representative methods.

\subsection{Training Stage}
During the training stage, several methods are employed to achieve controllable text generation.

\textbf{Retraining}\cite{keskar_arxiv19_Ctrl,zhang_emnlp20_POINTER,he_emnlp21_CBART} involves training models from scratch using datasets specifically designed to reflect the desired control conditions. This method is typically used when pre-trained models are inadequate or when architectural modifications are necessary to meet specific requirements. Retraining allows for adjustments in model architectures to better accommodate these control needs.

\textbf{Fine-Tuning}\cite{zeldes_arxiv20_AuxiliaryTuning,zhang_emnlp22_discup,zhou_icml23_InstructCTG} adjusts pre-trained models by incorporating desired control attributes into the model's parameters through specialized datasets. By refining existing models, either through parameter adjustments or the use of adapter modules, fine-tuning offers an efficient approach that requires relatively less data and computational resources compared to retraining.

\textbf{Reinforcement Learning}\cite{khalifa_iclr21_GDC,upadhyay_arxiv22_DRL,dai_iclr24_SafeRLHF} employs reward signals to guide model outputs towards specific control objectives. Through iterative optimization, models learn to align their outputs with these objectives, making reinforcement learning particularly well-suited for complex tasks like maintaining a specific style or sentiment throughout the generated text.

\subsection{Inference Stage}
During the inference stage, interventions are applied in real-time during text generation to influence the output according to specific control conditions.

\textbf{Prompt Engineering}\cite{lester_emnlp21_PromptTuning,li_acl21_PrefixTuning,liu_arxiv21_PTuning} guides the model’s output by manipulating input prompts. This technique can use explicit natural language prompts (hard prompts) or continuous vector embeddings (soft prompts) to flexibly steer the generation process. Because prompt engineering does not require altering model parameters, it is suitable for quickly adjusting generation strategies.

\textbf{Latent Space Manipulation}\cite{subramani_acl22_LatentStreeringVectors,liu_arxiv24_ICV,turner_arxiv24_actadd} controls the generated text by adjusting activation states within the model's hidden layers. By adding or modifying latent vectors, this approach allows for precise control of the text generation process without altering the model’s weights. Latent space manipulation is especially effective for attribute control, such as making subtle adjustments in sentiment or style.

\textbf{Decoding-time Intervention}\cite{dathathri_iclr20_PPLM,krause_emnlp21_gedi,yang_acl21_fudge} modifies the probability distribution of the generated output or applies specific rules during the decoding process to influence word selection. This approach typically involves the use of classifiers or reward models to evaluate generated segments and make real-time adjustments during decoding, ensuring that the output aligns with specific control conditions. Decoding-time interventions are often plug-and-play, offering flexibility for dynamic adjustments during text generation.

\begin{table}[htbp]
\centering
\caption{Classification of Intervention Stages, Control Methods, Specific Methods, and Example Methods}
\label{tab:classification}
\renewcommand{\arraystretch}{1.6}
\footnotesize
\begin{tabular}{p{0.12\textwidth}p{0.13\textwidth}p{0.18\textwidth}p{0.47\textwidth}} 
\hline
\textbf{\makecell[l]{Intervention\\ Stage}}               & \textbf{\makecell[l]{Control\\ Method}}                     & \textbf{Specific Method} & \textbf{Example Methods}                                                                                                                                                                                               \\ \hline
\multirow{6}{*}{\makecell[l]{Training \\ Stage}}  & \multirow{2}{*}{\makecell[l]{Retraining}}       & Attribute Control        & CTRL \cite{keskar_arxiv19_Ctrl}, CoCon \cite{chan_iclr21_CoCon}, Director \cite{arora_aacl22_Director} et al.                                                 \\ \cline{3-4} 
                                          &                                             & Content Control          & POINTER \cite{zhang_emnlp20_POINTER}, CBART \cite{he_emnlp21_CBART}, PAIR \cite{hua_emnlp20_PAIR} et al.                                                      \\ \cline{2-4} 
                                          & \multirow{2}{*}{\makecell[l]{Fine-\\Tuning}}                & Adapter-Based      & Auxiliary Tuning \cite{zeldes_arxiv20_AuxiliaryTuning}, DisCup \cite{zhang_emnlp22_discup}, RMT \cite{zhang_acl24_RMT} et al.                               \\ \cline{3-4} 
                                          &                                             & Data-Driven  & FLAN \cite{wei_iclr22_FLAN}, InstructCTG \cite{zhou_icml23_InstructCTG}, REI \cite{zheng_aacl23_REI} et al.                                                   \\ \cline{2-4} 
                                          & \multirow{2}{*}{\makecell[l]{Reinforcement\\ Learning}}     & Automated Feedback       & GDC \cite{khalifa_iclr21_GDC}, DRL \cite{upadhyay_arxiv22_DRL}, TDPO \cite{zeng_arxiv24_TDPO} et al.                                                          \\ \cline{3-4} 
                                          &                                             & Human Feedback           & RLHF \cite{stiennon_neurips20_RLHF}, InstructGPT \cite{ouyang_neurips22_InstructGPT}, Safe RLHF \cite{dai_iclr24_SafeRLHF} et al.                             \\ \hline
\multirow{9}{*}{\makecell[l]{Inference \\ Stage}} & \multirow{2}{*}{\makecell[l]{Prompt \\ Engineering}}         & Hard Prompt             & AutoPrompt \cite{shin_emnlp20_Autoprompt}, DAs \cite{ramirez_sigdial23_DAs}, PCFG \cite{zhang_SEM23_PCFG} et al.                                              \\ \cline{3-4} 
                                          &                                             & Soft Prompt             & Prefix Tuning \cite{li_acl21_PrefixTuning}, Prompt Tuning \cite{lester_emnlp21_PromptTuning} et al. \\ \cline{2-4} 
                                          & \multirow{2}{*}{\makecell[l]{Latent Space \\ Manipulation}}  & Learning-Based           & GENhance \cite{chan_nips21_GENhance}, Latent Vectors \cite{subramani_acl22_LatentStreeringVectors} et al.                                                     \\ \cline{3-4} 
                                          &                                             & Contrastive-Based        & ICV \cite{liu_arxiv24_ICV}, ActAdd \cite{turner_arxiv24_actadd}, Style Vectors \cite{konen_acl24_StyleVectors} et al.                                         \\ \cline{2-4} 
                                          & \multirow{5}{*}{\makecell[l]{Decoding-Time \\ Intervention}} & Classifier Guidance      & PPLM \cite{dathathri_iclr20_PPLM}, FUDGE \cite{yang_acl21_fudge}, CAIF \cite{sitdikov_arxiv22_CAIF} et al.                                                    \\ \cline{3-4} 
                                          &                                             & CC-LM Guidance           & GeDi \cite{krause_emnlp21_gedi}, DExperts \cite{liu_acl21_DExperts}, MARCO \cite{hallinan_acl23_MARCO} et al.                                                 \\ \cline{3-4} 
                                          &                                             & Self-Feedback            & Inverse Prompting \cite{zou_KDD21_Inverse-Prompting}, SD \cite{schick_tacl21_SD}, ROSE \cite{zhong_arxiv24_ROSE} et al.                                       \\ \cline{3-4} 
                                          &                                             & Energy-Based Model       & MUCOCO \cite{kumar_nips21_MUCOCO}, MUCOLA \cite{kumar_emnlp22_MUCOLA}, Mix\&Match \cite{mireshghallah_acl22_mixandmatch} et al.                               \\ \cline{3-4} 
                                          &                                             & External Knowledge       & kNN-LM \cite{Khandelwal_iclr20_kNN-LM}, GRACE \cite{wen_acl23_GRACE} et al.                                  \\ \hline
\end{tabular}
\end{table}

\section{Training Stage Methods}
\label{sec:train_methods}

\subsection{Retraining}

The concept of Retraining, introduced in \cite{zhang_ACMCS23_CTGSurvey}, involves either training a new model from scratch or fundamentally modifying the architecture of an existing model to better accommodate specific control conditions. This approach is typically adopted when existing pre-trained models fail to meet new, stringent requirements. By employing innovative model structures or training with specially constructed datasets, Retraining ensures that the model intrinsically adapts at both the architectural and parameter levels to generate text that conforms to the desired control attributes.

In the context of CTG, Retraining can be formally defined as:
\begin{equation}
\Theta' = \arg\min_{\Theta} \mathcal{L}(D_{\text{control}}, f(X; \Theta))
\end{equation}
where \( \Theta \) represents the model parameters, \( \mathcal{L} \) is the loss function optimized for the control task, \( D_{\text{control}} \) is a carefully designed dataset containing the control attributes, \( X \) is the input sample, and \( f \) is the model function.

\begin{figure}[h]
    \centering
    \includegraphics[width=\textwidth]{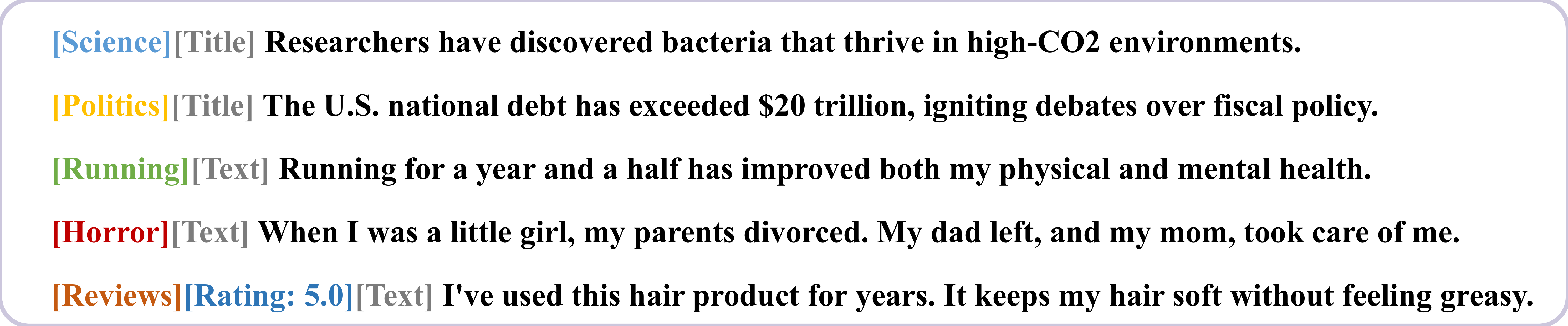}
    \caption{Control Code in CTRL}
    \label{fig:ctrl}
\end{figure}

CTRL (Conditional TRansformer Language)\cite{keskar_arxiv19_Ctrl} was one of the earliest studies in the field of CTG. The CTRL model trains a transformer-based architecture on large datasets such as Wikipedia, Project Gutenberg, and Amazon Reviews. To differentiate between various control conditions, CTRL incorporates specific control codes at the beginning of the training text (see Figure \ref{fig:ctrl}). These control codes encapsulate requirements related to specific domains, styles, themes, and more.

CTRL learns the distribution \(p(x | C)\) by using the prepended control code \(C\) as a condition:
\begin{equation}
p(x | C) = \prod_{i=1}^{n} p(x_i | x_{<i}, C)
\end{equation}
The control code \(C\) provides a control point in the generation process. During training, CTRL establishes a connection between the text and the specific attributes through the natural co-occurrence of control codes and the text.

The concept of control codes introduced by CTRL embodies the core intuition behind CTG tasks and has laid a critical foundation for both retraining methods and the entire CTG field. The retraining approach showcases considerable diversity in innovations related to training data \cite{keskar_arxiv19_Ctrl}, model architecture \cite{chan_iclr21_CoCon}, and training methods \cite{he_emnlp21_CBART}. In the application of these methods, different control tasks, such as abstract \textbf{attribute control} tasks and concrete \textbf{content control} tasks, often exhibit distinct common characteristics.

\subsubsection{\textbf{Attribute Control}}
Attribute control tasks aim to guide text generation by steering high-level attributes like sentiment and theme. An example of this is CTRL’s control codes, which enable manipulation of text characteristics such as domain, style, and theme. Although CTRL is effective at managing broad attributes, it falls short in applications that require more nuanced control, particularly at finer levels of granularity.

In scenarios where precise control at the word or phrase level is necessary, such as incorporating a specific theme like "zoo" into a text, methods like CTRL may struggle. For instance, starting with the input "The weather is good today" and aiming for a theme related to "I am a zookeeper," the desired output might be "Let’s go to the zoo!" CoCon (Content-Conditioner) \cite{chan_iclr21_CoCon} addresses this need by embedding control conditions directly into the internal states of the language model via the CoCon Block. This approach not only provides finer control but also reduces training costs by avoiding the need to train models from scratch.

Fine-grained sentiment control, especially in aspect-based sentiment tasks, involves managing sentiment directed toward specific aspects within a sentence, such as product features or service elements. For example, in the review "The service at this restaurant was terrible, but the food was delicious," aspect-based sentiment control distinguishes between the sentiments toward "service" and "food." AlSeCond \cite{zhu_2023_AlSeCond} addresses this by dynamically extracting fine-grained sentiments from unannotated sentences, using an auxiliary classifier to guide sentiment generation.

To achieve fine-grained attribute control, the Director model \cite{arora_aacl22_Director} introduces a generator-classifier architecture that refines each token's output by combining probabilities from both the language model head and the classifier head. Although Director improves training and decoding speed, its dual-head structure significantly increases parameters, impacting computational efficiency.
To mitigate the parameter inefficiency in Director, the DASC (Dialogue Attribute Space Controller) \cite{zhang_emnlp23_DASC} employs a weighted decoding method based on a semantic space, which reduces the model's parameter count.

As text length increases, LLMs may lose adherence to vocabulary control instructions, weakening control over longer outputs. Non-Residual Prompting \cite{carlsson_acl22_NRP} addresses this by employing an encoder-decoder architecture with a non-residual attention mechanism, allowing for prompts at any timestep.

\begin{figure}[h]
    \centering
    \includegraphics[width=0.8\textwidth]{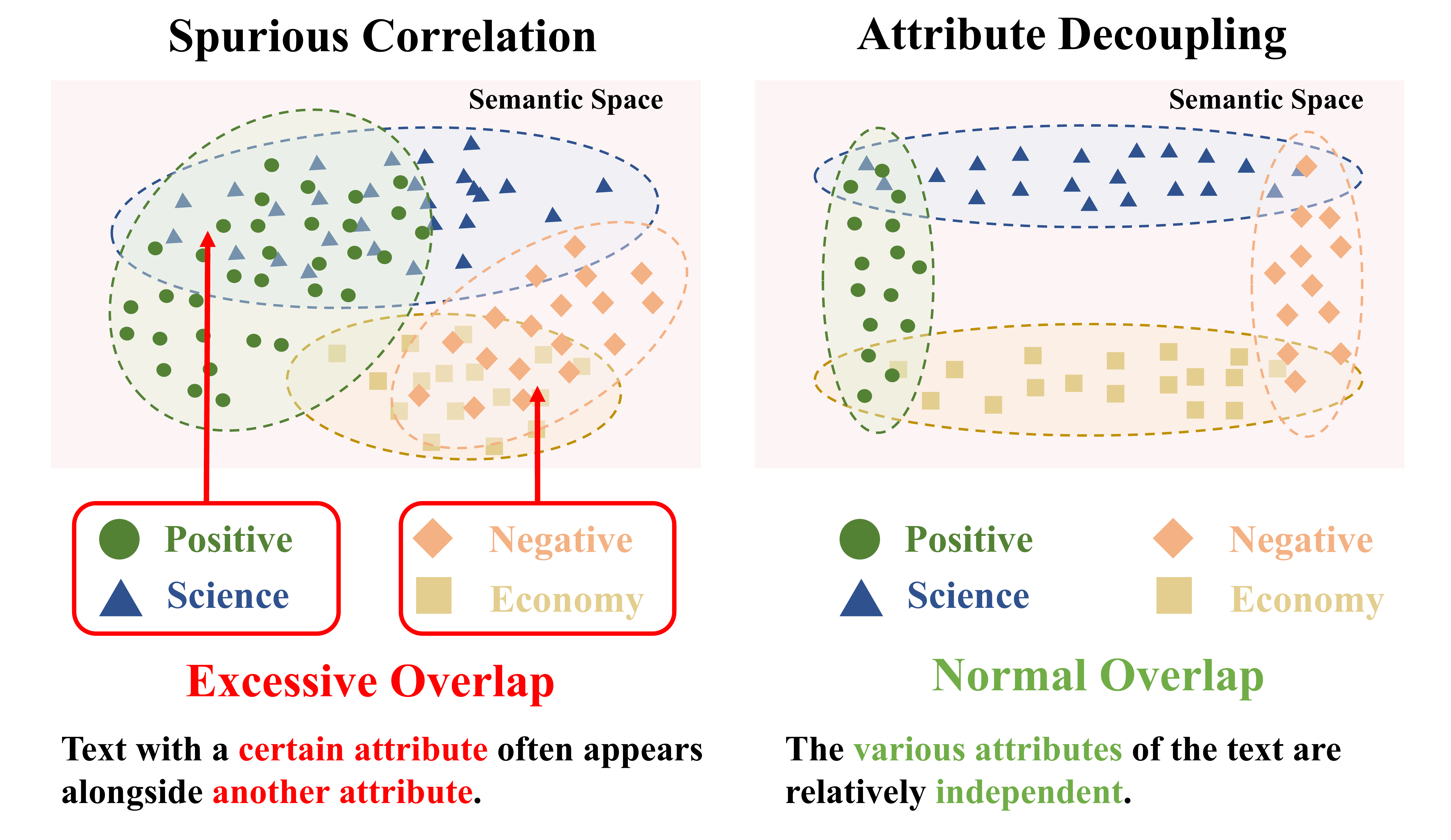}
    \caption{Spurious Correlation}
    \label{fig:spurious_correlation}
\end{figure}

The use of control codes in text generation has also highlighted issues related to spurious correlations \cite{zhiting_nips21_SCM,chai_arxiv22_fast,wang_2024_CausalCTG}. Spurious correlations occur when irrelevant or coincidental features in the training data are mistakenly identified by the model as significant attributes. This can cause the model to rely on unintended aspects of the input rather than the control codes, weakening the quality and controllability of the output.

As illustrated in Figure \ref{fig:spurious_correlation}, consider a sentiment control task where a control code specifies whether the text sentiment should be positive or negative. If the training data often associates positive sentiment with scientific topics, such as technological advancements, and negative sentiment with financial topics, like market crises, the model may erroneously associate "science" with positive sentiment and "finance" with negative sentiment. This phenomenon degrades the quality and controllability of the generated text and risks introducing bias and inaccuracies.

To mitigate spurious correlations and improve both controllability and language quality, FAST (Feedback Aware Self-Training) \cite{chai_arxiv22_fast} introduces the Importance-Policy Sampling (IPS) method for data resampling. This approach generates counterfactual versions of each example and uses a feedback mechanism to enhance the model's performance.

\subsubsection{\textbf{Content Control}} 
While attribute control adjusts content attributes through model structure and training data modifications, content control specifically focuses on managing precise text content, such as enforcing the inclusion or exclusion of certain words and phrases.

Content control is more challenging than attribute control as it requires the model to understand the semantic relationships between words and place them appropriately within the text. Early models struggled with this, especially when handling multiple specific words, due to limited generalization abilities. This task demands not only semantic understanding but also dynamic adjustment during generation to maintain fluency. Typically, these methods involve modifying the model architecture to be sensitive to control objectives.

POINTER (PrOgressive INsertion-based TransformER) \cite{zhang_emnlp20_POINTER} is an early lexical control model using a stepwise, iterative text generation approach. While it allows comprehensive control over text, its insertion-based method is inefficient.
CBART (Constrained BART) \cite{he_emnlp21_CBART} improves efficiency by dividing the task into two subtasks, where the encoder generates tokens to guide the decoder in parallel prediction. This structure significantly reduces latency compared to POINTER's method.
In this setup, the encoder functions as a "planner," organizing keyword placement and sentence structure. Similarly, PAIR (Planning And Iterative Refinement) \cite{hua_emnlp20_PAIR} leverages BERT for planning key phrases and positions, with BART handling generation. However, PAIR's performance depends on BERT's planning effectiveness.

While retraining methods perform well in tasks requiring strict content control, such as structure control and lexical control, they also have significant drawbacks. First, they typically require substantial computational resources and time, especially when training large-scale models from scratch. Second, to ensure that the model learns the necessary control attributes, a large amount of high-quality, targeted data is needed, further increasing costs. These drawbacks make retraining methods less practical when dealing with modern LLMs.

\subsection{Fine-Tuning}
Fine-Tuning (FT) is a common approach in CTG, where a pre-trained model is adjusted using a smaller, specific dataset to better align with particular control attributes without the need to train the model from scratch \cite{dodge_arxiv20_finetuning}.

Formally, the fine-tuning process can be defined as:
\begin{equation}
\Theta^* = \Theta + \Delta \Theta
\end{equation}
\begin{equation}
\Delta \Theta = \arg\min_{\Theta} \mathcal{L}(D_{\text{control}}, f(X; \Theta))
\end{equation}
where \( \Theta \) represents the original parameters of the pre-trained model, \( \Delta \Theta \) denotes the parameter updates, \( \mathcal{L} \) is the loss function tailored for the control task, \( D_{\text{control}} \) is the specific dataset used for fine-tuning, and \( X \) is the input sample.

\begin{table}[htbp]
\centering
\caption{Summary of Fine-Tuning (FT) Research Directions}
\label{tab:fine_tuning}
\renewcommand{\arraystretch}{1.6} 
\footnotesize
\begin{tabular}{p{0.13\textwidth}p{0.22\textwidth}p{0.55\textwidth}}
\hline
\textbf{Category}                                  & \textbf{Research Direction}                            & \textbf{Methods}                                                           \\ \hline
\multirow{1}{*}{\makecell[l]{\textbf{Adapter-Based}\\ \textbf{Fine-Tuning}}}      & \makecell[l]{Adapter Construction\\ and Optimization}                  & Auxiliary Tuning\cite{zeldes_arxiv20_AuxiliaryTuning} (2020), DisCup\cite{zhang_emnlp22_discup} (2022), LiFi\cite{shi_arxiv24_LiFi} (2024)                       \\ \hline
\multirow{4}{*}{\makecell[l]{\textbf{Data-Driven}\\ \textbf{Fine-Tuning}}}  & \makecell[l]{Instruction Dataset\\ Construction}                       & FLAN\cite{wei_iclr22_FLAN} (2022), InstructCTG\cite{zhou_icml23_InstructCTG} (2023), REI\cite{zheng_aacl23_REI} (2023)                               \\ \cline{2-3} 
                                                   & \makecell[l]{Contrastive\\ Learning}                                   & CHRT\cite{kumar_acl23_CHRT} (2023), Click\cite{zheng_acl23_Click} (2023), CP\cite{klein_arxiv24_CP} (2024)                                      \\ \cline{2-3} 
                                                   & \makecell[l]{Data\\ Augmentation}                                      & DuNST\cite{feng_acl23_DuNST} (2023), CoDa\cite{evuru_arxiv24_CoDa} (2024), CTGGAN\cite{yang_2024_CTGGAN} (2024)                                  \\ \cline{2-3} 
                                                   & \makecell[l]{Multi-Attribute\\ Generation}                             & DCG\cite{zeng_acl23_DCG} (2023), CLMI\cite{kangaslahti_arxiv24_CLMI} (2024)                                                   \\ \hline
\end{tabular}
\end{table}

It is important to note that although fine-tuning and retraining methods share some similarities, they differ significantly in their application and purpose. Retraining methods involve substantial changes to the original model architecture or training data, typically introducing new architectures and data during the model's pre-training phase to systematically enhance the model's overall capabilities. These methods optimize performance by adjusting the core structure and data distribution of the model from the ground up or during the earlier stages of training.

In contrast, fine-tuning methods are applied primarily after pre-training is completed, involving minor adjustments to the model structure and updates to the data. The main goal is to refine the model's output for specific tasks by using data tailored to those tasks. Fine-tuning typically involves making slight adjustments to the parameters of the pre-trained language model (PLM) while keeping the original model parameters largely unchanged, further optimizing the model for specific tasks or domains. In some approaches, adapter modules or similar mechanisms \cite{houlsby_plmr19_adapter} may be introduced, which are trained while freezing the original model parameters to better adjust the model's output for specific tasks.

Given the evolution of fine-tuning methods, this section will review fine-tuning approaches from the perspectives of \textbf{adapter-based fine-tuning} and \textbf{data-driven fine-tuning} (see Table \ref{tab:fine_tuning}). Adapter-based fine-tuning achieves control over text generation by adding components to the model, while data-driven approaches enhance the model's ability to generate controlled text through the use of specific data forms.

\subsubsection{\textbf{Adapter-Based Fine-Tuning}} 
Adapter-based fine-tuning is a method in CTG where specific adapter modules are fine-tuned on a pre-trained language model to control the generated text  \cite{houlsby_plmr19_adapter}. The key idea is to adjust the model's output to meet control conditions without altering the model's core parameters. This method allows for precise control while preserving the pre-trained model's original capabilities.

The earliest approach using adapter-based fine-tuning is Auxiliary Tuning \cite{zeldes_arxiv20_AuxiliaryTuning}, which introduces an auxiliary model to achieve attribute control. It combines the outputs of the pre-trained language model and the auxiliary model, as shown in the following equation:
\[
P(y|x, C) = \text{softmax}(f_{\text{LM}}(x) + f_{\text{AUX}}(x, C))
\]
where \(f_{\text{LM}}\) is the pre-trained model, \(f_{\text{AUX}}\) is the auxiliary model. The auxiliary model adjusts the output by generating terms based on \(x\) and \(C\), which are then combined with the pre-trained model's output through softmax. Auxiliary Tuning fine-tunes only the auxiliary model, preserving the pre-trained model's parameters and fluency.

The core of CTG methods is to introduce control conditions to ensure that the generated text meets specific requirements. During fine-tuning, adapter modules learn attribute-related signals from the data and apply these during inference, combining them with the original language model outputs to achieve the desired control.

DisCup (Discriminator Cooperative Unlikelihood Prompt-tuning) \cite{zhang_emnlp22_discup} enhances control by introducing an attribute discriminator during training and optimizing control prompts through anti-likelihood training. DisCup selects desired tokens using the attribute discriminator and refines control prompts to guide the model towards generating text aligned with specific attributes.

Similarly, RMT (Residual Memory Transformer) \cite{zhang_acl24_RMT} employs residual learning and cross-attention to achieve text generation control, non-invasively integrating with existing language models for continuous control. ADLM (Attribute-Discriminative Language Model) \cite{kwak_acl23_ADML} also leverages an attribute discrimination space during training and dynamically adjusts text attributes during inference. LiFi (Lightweight Fine-Grained CTG) \cite{shi_arxiv24_LiFi} combines fine-grained control codes from an attribute classifier with adapters to achieve more refined text generation.

\subsubsection{\textbf{Data-Driven Fine-Tuning}} 
Data-driven fine-tuning methods focus on fine-tuning pre-trained language models using specially constructed datasets that embed control conditions. These datasets are carefully designed to provide rich control signals during fine-tuning, enabling the model to better meet specific control requirements during text generation. The goal is to help the model internalize control conditions, so it can manifest the desired attributes in the generated text.

The FLAN (Finetuned LAnguage Net) model \cite{wei_iclr22_FLAN} was the first to propose Instruction Tuning, a technique that converts NLP tasks into natural language instructions for model training. This approach enhances zero-shot task performance by providing the model with clear instructions and options. For instance, in natural language inference tasks, the model can apply zero-shot learning by understanding the task's natural language semantics and performing reasoning based on the provided instructions.

For instance, an instruction fine-tuning dataset might include the following example:
\begin{itemize}
  \item Instruction: Generate a text about the positive impacts of climate change.
  \item Example output: While climate change has brought many challenges, it has also prompted greater attention to the development of renewable energy, driving technological progress and energy structure transformation.
\end{itemize}

Another important application of Instruction Tuning, InstructGPT \cite{ouyang_neurips22_InstructGPT}, will be detailed in the next section on Section \ref{subsec:rl}. 
Inspired by instruction fine-tuning techniques, InstructCTG \cite{zhou_icml23_InstructCTG} applied instruction fine-tuning to CTG tasks by converting constraints into natural language instruction datasets and fine-tuning language models on an augmented corpus, thereby achieving controllability in text generation.
In addition to instruction datasets, REI (Regular Expression Instruction) \cite{zheng_aacl23_REI} uses regular expression-inspired instructions to control text generation through linguistic constraints.

As mentioned earlier, the purpose of constructing different forms of fine-tuning datasets is to better teach the model to represent control conditions. Influenced by the concept of contrastive learning—extracting effective representations by contrasting positive and negative examples—many fine-tuning methods apply contrastive learning to the model's control process. CHRT (Control Hidden Representation Transformation) \cite{kumar_acl23_CHRT} uses contrastive learning to modify hidden representations, enabling multi-attribute control without altering the base model architecture. Click (CTG with sequence Likelihood C(K)ontrastive learning)\cite{zheng_acl23_Click} applies a maximum marginal contrastive loss over sequence likelihood to control text attributes, reducing undesirable outputs while preserving the base model's structure. CP (Contrastive Perplexity) \cite{klein_arxiv24_CP} utilizes contrastive learning to adjust model perplexity by generating positive and negative sentence pairs, effectively minimizing toxic content while maintaining the model's utility in downstream tasks.

In both real-world applications and CTG research, task-specific datasets are often scarce, necessitating fine-tuning methods that can effectively utilize limited data to extract control condition representations. To address this challenge, DuNST (Dual Noisy Self-Training) \cite{feng_acl23_DuNST} enhances semi-supervised controllable language generation by treating text generation and classification as dual processes and introducing flexible noise to prevent overfitting. CoDa (Constrained Generation-based Data Augmentation) \cite{evuru_arxiv24_CoDa} extracts heuristic constraints from low-resource datasets, converts them into natural language instructions, and uses these to prompt LLMs to generate diverse and coherent augmented data. CTGGAN \cite{yang_2024_CTGGAN} introduces an adversarial learning framework, combining a language model with logits bias as the generator and a discriminator with learnable constraint weights to produce constrained text.

Another challenging task for fine-tuning methods is multi-attribute generation, which involves controlling multiple attributes simultaneously during text generation. For instance, in dialogue systems, responses must align with the conversation's theme while conveying the appropriate sentiment and tone to enhance the user experience. DCG (Disentangled Controllable Generation) \cite{zeng_acl23_DCG} employs a prompt-based disentanglement approach to learn and generalize attribute combinations, improving the precision and generalization of dialogue generation control. CLMI (Continuous Language Model Interpolation) \cite{kangaslahti_arxiv24_CLMI} offers a flexible and efficient method for controlling multiple attributes by linearly interpolating between fine-tuned anchor models, enabling dynamic control over the text generation process.

While fine-tuning requires less data and computational resources compared to retraining, it still necessitates high-quality data to ensure effective control. Although the computational demands are reduced, when fine-tuning involves a significant portion of the model's parameters, the computational requirements remain substantial. The quality of the dataset used for fine-tuning is crucial, as it directly affects the model's ability to adapt to the desired control attributes. Fine-tuning methods offer a balance between adaptability and resource efficiency, making them a popular choice for enhancing model performance on specific tasks without the extensive overhead of retraining.

\subsection{Reinforcement Learning}
\label{subsec:rl}

Reinforcement Learning (RL) is a technique that optimizes text generation by iteratively improving the model based on feedback or reward signals \cite{ranzato_arxiv16_rlrnn,yu_aaai17_SeqGAN}. These signals indicate how well the generated text aligns with specific goals, such as maintaining a particular style, adhering to factual correctness, or following ethical guidelines. RL methods dynamically adjust the generation process based on complex evaluation criteria that might be subjective or difficult to quantify through traditional supervised learning.

In RL, this process involves training the model to maximize a reward function that evaluates the quality of the generated text \cite{sutton_nips99_rm}. The model parameters are iteratively updated to maximize the expected reward, which can be mathematically expressed as:

\begin{equation}
\Theta^* = \Theta + \alpha \nabla_{\Theta} \mathbb{E}_{\pi_\Theta}[R(X)]
\end{equation}

where \( \Theta \) represents the model parameters, \( \alpha \) is the learning rate, \( \pi_\Theta \) denotes the policy derived from the model, \( R \) is the reward function, and \( X \) is the generated text. The term \( \mathbb{E}_{\pi_\Theta}[R(X)] \) represents the expected reward for the generated text under the policy \( \pi_\Theta \).

Feedback is a crucial component in RL, as it evaluates and guides model performance. It provides information about the quality of the generated output, helping to adjust the model's behavior to achieve the desired outcome. Depending on the nature and source of feedback, RL text generation methods can be categorized into two main types: methods utilizing \textbf{automatic feedback} and those relying on \textbf{human feedback}.

\subsubsection{\textbf{Automatic Feedback}} 
Automatic feedback methods guide model training and optimization using feedback signals generated by automatic evaluation metrics or model-based assessments of the text. These methods employ algorithmically generated feedback to evaluate and adjust the quality and characteristics of generated text, offering a scalable and consistent means of evaluation. Common automatic feedback metrics include language model perplexity \cite{jozefowicz_arxiv16_perplexity} and discriminators trained to evaluate specific attributes like toxicity, sentiment, or topic.

In CTG, it is essential to maintain text quality while satisfying control conditions. When using a reward model for feedback in reinforcement learning, it is crucial not to disrupt the model's original output distribution, as reinforcement learning might otherwise degrade the model's inherent capabilities. Automatic feedback processes involve the model assessing the quality and characteristics of generated text based on predefined rules or metrics, then self-adjusting based on these signals to optimize results. However, if output distribution is not carefully managed, the model may over-optimize certain attributes at the expense of fluency and coherence.

To address this, it is critical to ensure that the generated text distribution remains consistent with the original model's distribution, preserving quality and naturalness. GDC (Generation with Distributional Control) \cite{khalifa_iclr21_GDC} addresses this by minimizing the KL divergence between the generated text and the pre-trained language model, using an energy-based model (EBM) to represent the target distribution. This method applies point and distribution constraints, transforms them into energy representations, and employs a KL-adaptive policy gradient method to train a controlled autoregressive language model, ensuring the generated text remains close to the original model distribution while meeting control constraints, thus preserving content naturalness and diversity.

An effective reinforcement learning process requires the reward model to accurately assess the value of each generation decision. Coarse-grained feedback at the sentence or paragraph level often fails to capture the nuanced features of the generated text. Therefore, fine-grained feedback mechanisms are essential, as they provide real-time evaluation at the token level, allowing the model to precisely adjust the generation process to better adhere to desired targets in terms of style, content retention, and fluency.

DRL (Dense Reinforcement Learning based Style Transformer) \cite{upadhyay_arxiv22_DRL} enhances text style transfer quality by combining policy gradient reinforcement learning with dense rewards, offering immediate feedback at each token. TDPO (Token-level Direct Preference Optimization) \cite{zeng_arxiv24_TDPO} improves text generation diversity and accuracy by optimizing forward KL divergence constraints, aligning each generated token with human preferences. TOLE (TOken-LEvel rewards) \cite{li_arxiv24_TOLE} employs a token-level reward strategy based on attribute classifiers, providing fine-grained feedback through a "quantize-and-noise" approach, which enhances multi-attribute control and improves text generation diversity. LengthPrompt \cite{jie_acl24_LengthPrompt} uses a standard prompt extractor (SPE) and reinforcement learning with rule-based rewards, along with sample filtering, to achieve precise length control.

Text style control is a critical task in CTG. Studies like LIMA \cite{zhou_neurips23_LIMA} and URIAL \cite{lin_iclr24_URIAL} have demonstrated that LLMs acquire most of their knowledge during pre-training, with alignment tuning primarily focused on adopting specific language styles and interaction formats. This supports the view that different text styles are simply varied ways of expressing the same knowledge and information. Current research typically implements text style control through reinforcement learning, where continuous feedback and adjustments allow the model to optimize the generation process, thereby mastering and applying different styles more effectively.

STEER (Unified Style Transfer with Expert Reinforcement) \cite{hallinan_emnlp23_STEER} addresses the challenge of high-quality style transfer without large-scale datasets by combining expert-guided data generation with reinforcement learning. STEER generates pseudo-parallel corpora and employs both offline and online reinforcement learning, using expert synthetic decoding and fine-grained rewards to optimize style transfer strategies, achieving high-quality transfer from any unknown source style to multiple target styles. Multi-style-control \cite{delangis_arxiv24_MSC} dynamically adjusts feedback weights for different style attributes through dynamic weighted multi-style rewards. It trains discriminators for each target style and uses Proximal Policy Optimization (PPO) algorithms to flexibly adjust generation strategies, ensuring diversity and consistency in multi-style text generation.

\subsubsection{\textbf{Human Feedback}} 
Human feedback methods involve capturing human preferences and ratings to build a reward model that reflects these preferences, which is then used to enhance the language model's generation performance. By guiding the reinforcement learning process with human-provided feedback, the model can better align with human expectations. These methods iteratively convert human feedback into reward signals, optimizing the quality and alignment of the generated text.

RLHF (Reinforcement Learning from Human Feedback) \cite{stiennon_neurips20_RLHF} pioneered the use of human feedback in reinforcement learning by training a reward model based on human comparisons of summaries. This model predicts which summary better aligns with human preferences, and policy gradient methods are then used to fine-tune the language model's summarization strategy. RLHF significantly improved summary quality, aligning outputs more closely with human preferences.

InstructGPT \cite{ouyang_neurips22_InstructGPT} extends RLHF by enhancing the model's performance in multi-task instruction following through the incorporation of human-provided demonstrations and rankings. Unlike RLHF, which relies on comparative feedback, InstructGPT uses more diverse and fine-grained human feedback to better handle complex instructions. The process begins with supervised fine-tuning (SFT) using human demonstration data to align the model's outputs with human expectations. Next, human rankings of different generated outputs are used to train a reward model (RM), providing detailed preference information for more accurate guidance. Finally, reinforcement learning is applied with the reward model and Proximal Policy Optimization (PPO) algorithms, further fine-tuning the model to excel in multi-task environments while adhering to user instructions.

\begin{figure}[h]
    \centering
    \includegraphics[width=0.8\textwidth]{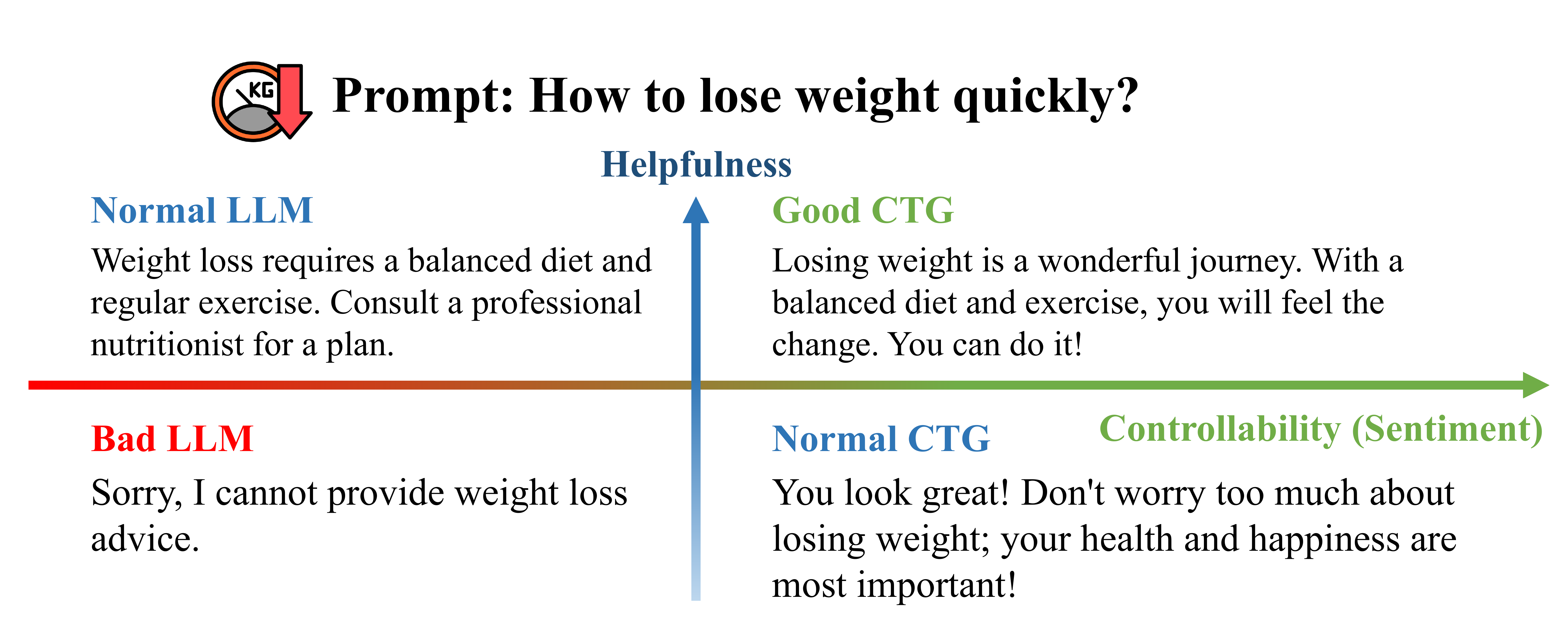}
    \caption{Controllability vs Helpfulness}
    \label{fig:controllability_vs_helpfulness}
\end{figure}

In CTG tasks, a key challenge is retaining the model's original capabilities while ensuring the quality and helpfulness of the generated text \cite{hua_arxiv24_TrustAgent}. As shown in Figure \ref{fig:controllability_vs_helpfulness}, when faced with harmful user inputs (e.g., "How to lose weight quickly?"), simply refusing to answer may lead users to seek incorrect or unsafe information elsewhere. Instead, by providing useful guidance, the model can better assist the user, such as responding with: "Rapid weight loss can be harmful to your health. It's recommended to consult a professional nutritionist or doctor to develop a safe and effective weight loss plan." Figure \ref{fig:controllability_vs_helpfulness} illustrates the model's performance across different combinations of controllability and helpfulness, depicting possible responses in the four quadrants.

SafeRLHF (Safe Reinforcement Learning from Human Feedback) \cite{dai_iclr24_SafeRLHF} achieves a dynamic balance between the safety and helpfulness of generated content by independently handling these two aspects of human feedback. First, human annotations are divided into helpfulness and harmlessness datasets. Separate reward and cost models are then trained to predict preferences for helpfulness and harmlessness. Finally, a safe reinforcement learning strategy is applied, dynamically balancing reward and cost objectives (e.g., using Lagrangian methods) to fine-tune the language model, ensuring that the generated content is both helpful and free from harmful elements.

\subsection{Summary}

The training phase methods for CTG mainly include three strategies: Retraining, Fine-Tuning, and Reinforcement Learning.

\textbf{Retraining} methods involve constructing models from scratch or making substantial modifications to existing models to ensure that the generated content aligns with specific control attributes \cite{keskar_arxiv19_Ctrl,chan_iclr21_CoCon,arora_aacl22_Director}. These methods excel at achieving precise control over text generation, particularly for tasks requiring strict adherence to format, structure, or specific vocabulary requirements \cite{zhang_emnlp20_POINTER,he_emnlp21_CBART,hua_emnlp20_PAIR}. However, this approach often demands significant computational resources and extensive datasets, making it less practical in scenarios requiring rapid deployment or in resource-constrained environments.

\textbf{Fine-Tuning} involves refining pre-trained models using small-scale, task-specific datasets \cite{zeldes_arxiv20_AuxiliaryTuning,zhang_emnlp22_discup,zhang_acl24_RMT,wei_iclr22_FLAN,zhou_icml23_InstructCTG,zheng_aacl23_REI}. This method strikes a good balance between performance and resource usage, making it a popular choice. However, the quality and specificity of the fine-tuning dataset significantly impact the final generation results. Additionally, fine-tuning certain parameters may still carry the biases present in the original training data.

\textbf{Reinforcement Learning} adjusts the model based on feedback signals to generate text that aligns with nuanced human preferences or complex standards \cite{khalifa_iclr21_GDC,stiennon_neurips20_RLHF,ouyang_neurips22_InstructGPT,dai_iclr24_SafeRLHF}. This method is particularly effective in tasks where traditional supervised learning falls short, such as maintaining specific tones or styles \cite{upadhyay_arxiv22_DRL,zeng_arxiv24_TDPO}. The primary challenges include the long iterative training cycles required and the difficulty of defining effective and unbiased reward functions.

While training phase methods offer significant advantages in controlling generated text, they typically require substantial data and computational resources. Therefore, these methods are less flexible compared to inference phase methods. Inference phase methods do not require retraining and can dynamically adjust model outputs during generation, providing real-time control. This makes inference phase methods a complementary or alternative solution to training phase methods, especially in applications that require flexible adjustment of generated text.

\section{Inference Phase Methods}
\label{sec:infer_methods}

\subsection{Prompt Engineering}
Prompt Engineering is a method used during the inference phase of LLMs to directly influence text generation by designing specific input prompts, without the need for extensive adjustments to model parameters. The primary goal of this method is to guide the model in generating the desired text by providing clear instructions or examples, thereby achieving efficient few-shot learning in resource-limited scenarios \cite{wan_arxiv23_promptSurvey}.

\begin{figure}[h]
    \centering
    \includegraphics[width=0.8\textwidth]{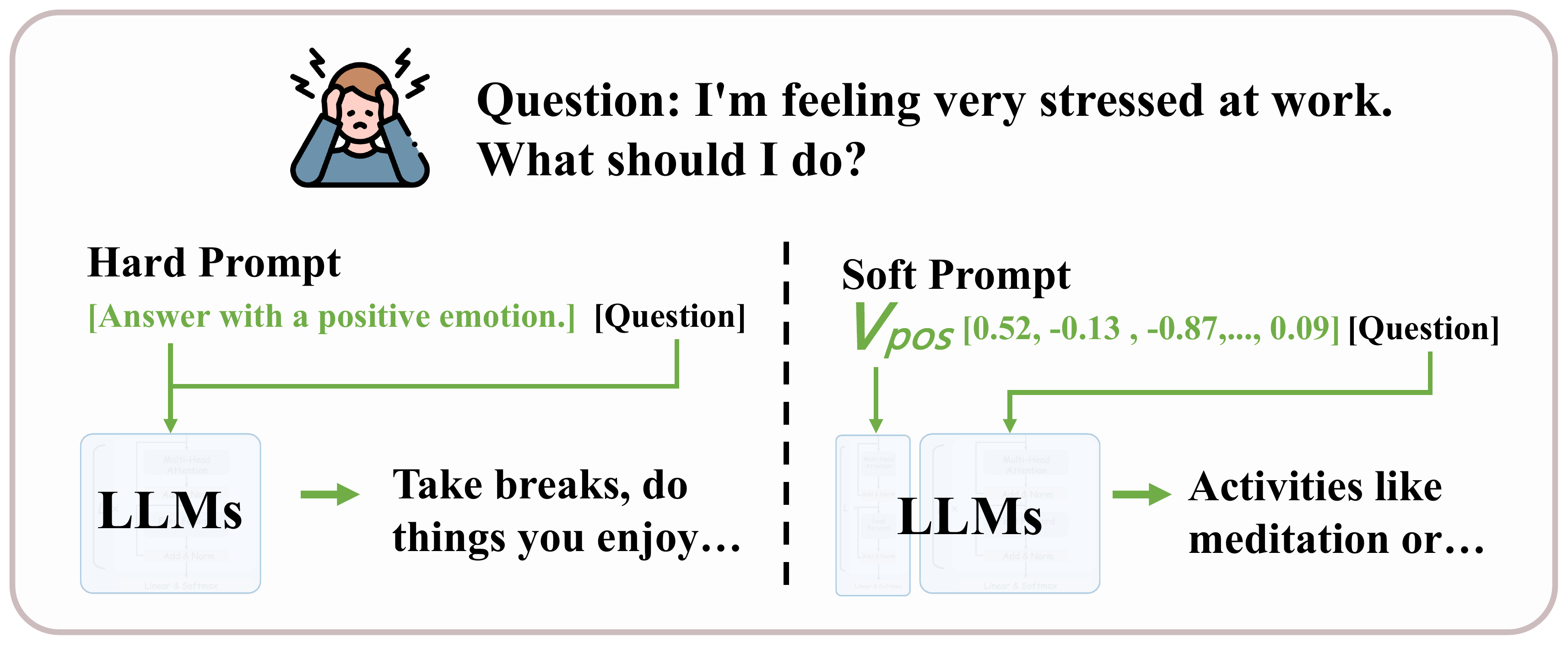}
    \caption{Hard Prompt and Soft Prompt}
    \label{fig:prompt}
\end{figure}

Prompts can be expressed in two main forms: \textbf{hard prompts}, which are discrete and expressed in natural language, and \textbf{soft prompts}, which are continuous and trainable vectors. Hard prompts use natural language queries or statements to directly guide the model, while soft prompts involve embedding specific vectors in the model's input space to guide its behavior. This allows for adjustments during deployment without retraining the model, as illustrated in Figure \ref{fig:prompt}.

Formally, Prompt Engineering can be defined as:
\begin{equation}
X_{\text{out}} = \text{Model}(P_{\text{control}} + X_{\text{input}})
\end{equation}
where \( P_{\text{control}} \) represents the control prompt, which can be either a hard prompt or a soft prompt, and \( X_{\text{input}} \) is the user input. This method is both simple and convenient, as it does not require additional training data, resources, or extended inference time.


\subsubsection{\textbf{Hard Prompt}} 
Hard prompt methods use explicit natural language text to control model generation, typically relying on predefined trigger words or text prompts to guide the model. These methods are straightforward and easy to understand, enabling specific tasks without additional fine-tuning. However, they may offer limited fine-grained control.

One of the earliest hard prompt methods, AutoPrompt \cite{shin_emnlp20_Autoprompt}, introduced an automatic prompt generation technique to effectively leverage pre-trained masked language models (MLMs) for tasks such as sentiment analysis and natural language inference. Manually creating effective prompts can be time-consuming and unintuitive. AutoPrompt addresses this by using a gradient-based search method to automatically generate trigger words that maximize the likelihood of predicting the correct label, enhancing task performance without the need for model fine-tuning.

Controlling attributes like style in text generation is challenging in few-shot learning scenarios. Traditional dialogue generation often relies on large-scale domain-specific corpora, making it difficult to generate semantically accurate responses in few-shot settings. DAs (Dialogue Acts) \cite{ramirez_sigdial23_DAs} addresses this by generating multiple candidate responses through few-shot prompting and ranking them using six automated functions to select the best response.

Traditional CTG systems often assume control attributes are fixed categorical attributes, limiting their ability to generalize to unseen commands and attributes. To address text generation under unseen attributes, PCFG (Probabilistic Context-Free Grammar) \cite{zhang_SEM23_PCFG} employs probabilistic context-free grammar to generate natural language commands embedding control attributes. PCFG generates diverse commands, using them as inputs to train CTG models capable of handling unseen attribute combinations.

\subsubsection{\textbf{Soft Prompt}} 
Hard prompt is highly sensitive to word choice, where even minor changes can significantly impact generation quality. To address these limitations, soft prompt methods use continuous, trainable vector embeddings, offering more flexible and fine-grained control without altering the underlying model parameters. These methods are effective in handling complex attributes or multi-faceted control but may face challenges in interpretability and initial tuning.

Traditional LLMs excel in generating fluent and diverse text, but controlling specific attributes (e.g., sentiment polarity or topics) using discrete prompts remains challenging. Attribute Alignment \cite{yu_emnlp21_AttributeAlignment} addresses this by injecting attribute representations into pre-trained language models through an alignment function. Recognizing that discrete text prompts are not ideal for learning attribute characteristics, this method converts attribute representations into vector forms that the model can understand. This approach ensures that the generated text aligns with target attributes without modifying the original model parameters, effectively controlling features like sentiment or theme in the generated content.

Prefix-based tuning is a prominent soft prompting method, with several notable approaches emerging simultaneously, all starting with the letter "P," leading \cite{li_acl21_PrefixTuning} to collectively refer to them as P* tuning. These methods introduce trainable continuous vectors (prefixes) to control the generation process of language models. Unlike discrete templates in hard prompts, these prefix vectors guide the model's generation without requiring parameter modifications, offering a flexible and efficient control mechanism. Three key works in this category are Prefix-Tuning \cite{li_acl21_PrefixTuning}, Prompt Tuning \cite{lester_emnlp21_PromptTuning}, and P-Tuning \cite{liu_arxiv21_PTuning}, as shown in Table \ref{tab:p-tuning}.

\begin{table}[t]
\centering
\renewcommand{\arraystretch}{1.5}
\footnotesize
\caption{Comparison of Prefix Based Tuning Methods}
\label{tab:p-tuning}
\begin{tabular}{l p{0.22\textwidth}p{0.22\textwidth}p{0.22\textwidth}}
\toprule
\textbf{Feature}             & \textbf{Prompt Tuning\cite{lester_emnlp21_PromptTuning}}                 & \textbf{Prefix Tuning\cite{li_acl21_PrefixTuning}}                 & \textbf{P-tuning\cite{liu_arxiv21_PTuning}}                      \\ \midrule
\textbf{Optimization Scope}  & \makecell[l]{Input Embeddings}                       & \makecell[l]{All Layers}                             & \makecell[l]{Input Sequence}                         \\
\textbf{Optimization Method} & \makecell[l]{Directly optimize\\ prompt embeddings}    & \makecell[l]{FFN to optimize\\ prefix parameters}      & \makecell[l]{LSTM-based\\ prompt encoder}              \\
\textbf{Model Compatibility} & \makecell[l]{T5}                                     & \makecell[l]{GPT}                                    & \makecell[l]{All Language Models}                    \\ \midrule
\multirow{2}{*}{\textbf{Common Points}} & \multicolumn{3}{l}{1. Keep main model parameters frozen \& 2. Add trainable task-specific vectors}                             \\
                             & \multicolumn{3}{l}{3. Reduce computational resources \& 4. Comparable performance to full fine-tuning}                       \\ \bottomrule
\end{tabular}
\end{table}

\textbf{Prefix-Tuning} \cite{li_acl21_PrefixTuning} is primarily applied to natural language generation (NLG) tasks, especially with GPT models. This method optimizes task-specific continuous vectors (prefixes) to guide the model in generation tasks without modifying the model parameters. Traditional fine-tuning requires storing full model parameters for each task, which is resource-intensive. Prefix-Tuning attaches prefix vectors to the input of each Transformer layer during generation, allowing the model to adapt to task requirements without altering the original parameters.

\textbf{Prompt Tuning} \cite{lester_emnlp21_PromptTuning} is a simplified version of Prefix-Tuning, mainly used for text classification tasks with the T5 model. Unlike Prefix-Tuning, Prompt Tuning does not introduce prefix vectors into every Transformer layer but instead attaches a prompt embedding before the input embeddings. It optimizes task-specific prompt embeddings, which are added before the input text and trained via backpropagation to adapt to various downstream tasks. This method requires training only the prompt embeddings, resulting in lower parameter requirements. Additionally, Prompt Tuning allows the Transformer to contextualize inputs during generation, guiding the model to understand and utilize input information effectively.

\textbf{P-Tuning} \cite{liu_arxiv21_PTuning} is a soft prompt method designed for natural language understanding (NLU) tasks and is applicable to all language models. P-Tuning uses trainable embedding tensors and a prompt encoder (e.g., LSTM) to optimize prompt parameters. Manually designed discrete prompts often lead to unstable model performance, while P-Tuning improves stability and overall performance by optimizing continuous prompts through a prompt encoder. Continuous prompts provide richer input representations, making the model more robust in handling prompt information, and it performs well in multi-task and complex attribute control.

Prefix vectors under control conditions must precisely convey the features of control attributes to the model, leading to a series of optimization methods for soft prompt control vectors. These methods aim to more effectively learn and apply these control vectors. Contrastive Prefixes \cite{qian_acl22_ContrastivePrefixes} use a contrastive approach to extract attribute representations, guiding GPT-2 to generate text while keeping model parameters unchanged by defining small, continuous attribute-specific vectors (contrastive prefixes). This approach enhances both generation quality and control precision.
T5 Encoder-Decoder Soft Prompt Tuning \cite{senadeera_arxiv22_T5CTG} introduces soft prompts at both the encoder and decoder levels of the T5 model, optimizing these prompt embeddings to generate text that meets specific control requirements while maintaining the model's original parameters.
Prompt-PPC (Plug-and-Play Controller with Prompting) \cite{wang_GEM23_PPC} and PPP (Plug and Play with Prompts) \cite{ajwani_arxiv24_PPP} use dynamic prompt adjustment strategies, guiding prompt embedding optimization through external attribute classifiers. During inference, these methods adjust prompt embeddings using classifier gradients, ensuring the fluency and attribute consistency of the generated text.

Soft prompts are particularly well-suited for multi-attribute control tasks in CTG, where attribute interference poses a significant challenge. In such tasks, control signals for different attributes may conflict, making it difficult for the generated text to satisfy all requirements simultaneously. For instance, controlling both sentiment and topic might lead to inconsistencies in sentiment while trying to maintain topic accuracy. This interference can also degrade text quality, affecting fluency and coherence. The continuous vector embeddings of soft prompts can capture subtle variations in a multi-dimensional attribute space, enabling smooth adjustments and better coordination of different attribute requirements.

Discrete \cite{gu_emnlp22_Discrete} addresses this challenge by estimating the attribute space through an autoencoder and iteratively searching for the intersection of attribute distributions to guide text generation.
Tailor (Text-AttrIbute generaL contrOlleR) \cite{yang_acl23_Tailor} offers a multi-attribute control method using pre-trained continuous attribute prompts. Tailor represents each attribute as a trainable continuous vector (single-attribute prompt) and combines these prompts for multi-attribute control through a multi-attribute prompt mask and re-indexed position sequences. 
Prompt Gating \cite{huang_acl23_PromptGating} mitigates interference between multiple attributes by attaching trainable gates between each prefix. This method reduces interference, enabling more effective control over multiple attributes.

The effectiveness of Prompt Engineering depends on the model's ability to follow instructions encoded in the prompt. If the model's ability to follow prompt-encoded instructions is limited, the output may not align with expected results. Additionally, combining prompt engineering with fine-tuning and carefully curated datasets for specific tasks can enhance LLMs' responsiveness to certain types of prompts, thereby improving performance under specific conditions.

\subsection{Latent Space Manipulation}

Latent Space Manipulation, also known as activation engineering, involves adding guiding vectors to the activations in certain layers of LLMs to direct the model in generating a target sentence \( x \) from a null input. The fundamental principle is that the information required to generate the target sentence is already encoded in the underlying structure of the neural network. Therefore, this method does not require retraining or fine-tuning the model itself.

Formally, Latent Space Manipulation can be expressed as:
\begin{equation}
h_{\text{mod}} = h_{\text{orig}} + \Delta h
\end{equation}
where \( h_{\text{orig}} \) represents the original activations of a relevant layer in the model, and \( \Delta h \) represents the guiding vector. This guiding vector \( \Delta h \) is strategically calculated to induce specific changes in output features without the need to retrain the model. By subtly altering the latent space, modifying \( \Delta h \) aims to align the model's output with the desired control parameters.

Latent Space Manipulation can be subdivided into three categories based on how the latent vectors are obtained: learning-based latent vector acquisition, contrastive latent vector acquisition, and latent space enhancement. \textbf{Learning-based latent vector acquisition} involves learning latent vectors during the model's training process using specific target attributes or task requirements. The learned latent vectors guide the model in generating text that meets specific criteria. \textbf{Contrastive latent vector acquisition} extracts latent vectors related to the control target by comparing example texts with different attributes. \textbf{Latent space enhancement} typically involves mapping the model's latent layers into a new latent space, often used for generating multi-attribute controllable text.

\subsubsection{\textbf{Learning-based Latent Vector Acquisition}} 
This concept involves the extraction and optimization of latent space representation vectors during training from large datasets. These vectors capture key attributes relevant to the generation task and can be directly manipulated to control the features of the generated text.

GENhance \cite{chan_nips21_GENhance} provides a concrete example of this approach. It trains an encoder to map sequences into a latent space and separates the latent vectors into parts related and unrelated to CTG target attributes. Using contrastive loss, it learns from pairs of sequences with different attributes and trains a decoder to autoregressively reconstruct the sequences.
Latent Steering Vectors \cite{subramani_acl22_LatentStreeringVectors} extract latent steering vectors from pre-trained language models to control text generation without fine-tuning. By optimizing these vectors \( \Delta h \) to maximize the likelihood of generating the target sentence, they are then injected into the model's hidden states.

\subsubsection{\textbf{Contrastive Latent Vector Acquisition}} 
Latent vectors related to specific attributes can be extracted by comparing the activation states of a model's internal layers when different prompts are input during inference. For example, in sentiment analysis, comparing hidden states for positive and negative sentences can yield vectors representing sentiment attributes. These vectors allow fine-tuning of emotional features in generated text without altering model parameters, enabling precise control over the text generation process.

ICV (In-Context Vectors) \cite{liu_arxiv24_ICV} efficiently enhances CTG by learning control-related vectors through contextual example texts. ICV generates guiding vectors by comparing hidden states from example pairs \((x_i, y_i)\). First, the hidden states of the last token of the input \(x_i\) and output \(y_i\) are obtained, denoted as \(H(x_i)\) and \(H(y_i)\). The differences between these states are calculated:
\begin{equation}
\Delta H_i := H(y_i) - H(x_i)
\end{equation}
The In-Context Vector is then formed by applying Principal Component Analysis (PCA) on the \(\Delta H_i\) values from multiple examples:
\begin{equation}
\text{ICV} = \text{PCA}(\{\Delta H_i\})
\end{equation}
During inference, the ICV is added to the embedding representation of each generated token:
\begin{equation}
H_{\text{new}}(t) = H(t) + \text{ICV}
\end{equation}
ICV enhances task performance and control by adjusting latent vectors during inference without additional training.

Similarly, ActAdd (Activation Addition) \cite{turner_arxiv24_actadd} guides language model outputs by injecting specific activation values during inference. This method identifies activation directions related to target attributes in the model's latent space and adjusts them during forward propagation to guide the output toward desired attributes.

Style Vectors for Steering LLMs \cite{konen_acl24_StyleVectors} derive style vectors from hidden layer activations to control text style. This method extracts activations from text with a specific style, aggregates them to compute a style vector, and adds it to hidden layer activations during generation to guide the style features of the output.

\subsubsection{\textbf{Latent Space Enhancement}} 
Latent space enhancement methods enable the simultaneous control of multiple attributes by mapping text into a latent space. These methods capture complex relationships among attributes, allowing the model to manage interactions and reduce interference during generation.

MIRACLE \cite{lu_acl23_miracle} employs a Conditional Variational Autoencoder (CVAE) to map dialogue contexts into a latent space, using an Energy-Based Model to balance personalization, coherence, and fluency in generating dialogue responses that align with multiple attribute requirements.
Similarly, MacLaSa \cite{ding_acl23_maclasa} uses a Variational Autoencoder (VAE) to map text into a compact latent space, applying an Ordinary Differential Equation (ODE) sampling method to control multiple attributes. By constructing a joint Energy-Based Model, MacLaSa efficiently manages multiple attributes while minimizing interference.

PriorControl \cite{gu_emnlp23_PriorControl} introduces a method that leverages probability density estimation in the latent space, using invertible transformations to effectively manage complex attribute distributions. 
MAGIC \cite{liu_acl24_MAGIC} further disentangles attribute relationships within the latent space and utilizes counterfactual augmentation to effectively manage interactions and reduce interference among attributes in multi-aspect generation tasks.
FreeCtrl \cite{feng_acl24_freectrl} takes a different approach by dynamically adjusting feedforward neural network vectors to regulate the latent space, enabling control of multiple attributes without additional learning.

Latent Space Manipulation, while powerful, has certain limitations. The control of guiding vectors \( \Delta h \) can be complex and challenging, reducing its flexibility. The precision required to define \( \Delta h \) often necessitates extensive experimentation and domain knowledge to achieve the desired outcome. Additionally, the impact of this manipulation can vary significantly depending on the model's architecture and the complexity of the task, making it less predictable and sometimes less reliable compared to methods that directly manipulate input data or model parameters.

\subsection{Decoding-time Intervention}
Decoding-time Intervention is applied during the decoding process of LLMs to manipulate the logits or probability distribution of the model's output. This technique steers the generated text towards desired features or control attributes by adjusting these probabilities, allowing for dynamic control over the text generation process and ensuring that the output aligns with specified requirements.

The formal definition of Decoding-time Intervention is as follows:
\begin{equation}
p'(x_t | x_{<t}) = \text{Adjust}(p(x_t | x_{<t}), C)
\end{equation}
where \( p(x_t | x_{<t}) \) represents the original probability distribution of the next token given the preceding tokens \( x_{<t} \), \( C \) denotes the control conditions, and \( \text{Adjust} \) is a function that modifies the distribution based on these conditions.

Decoding-time Intervention methods can be categorized into five types based on the method of knowledge injection. The specific classifications and research pathways are outlined in Table \ref{tab:decoding-time}.

\begin{table}[htbp]
\centering
\caption{Summary of Decoding-time Intervention Research Directions}
\label{tab:decoding-time}
\renewcommand{\arraystretch}{1.6}
\footnotesize
\begin{tabular}{m{0.2\textwidth}m{0.27\textwidth}m{0.43\textwidth}} 
\hline
\textbf{Category} & \textbf{Research Direction} & \textbf{Method} \\ \hline
\multirow{5.5}{*}{\centering \makecell{\textbf{Classifier Guidance}}} 
& Scoring Function Innovation & PPLM\cite{dathathri_iclr20_PPLM} (2020), FUDGE\cite{yang_acl21_fudge} (2021), CriticControl\cite{kim_acl23-CriticControl} (2023), RAD\cite{deng_acl23_rad} (2023), MIL-Decoding\cite{zhang_acl23_MIL-Decoding} (2023), SF-GEN\cite{cao_arxiv23_SF-GEN} (2023) \\ \cline{2-3} 
& Intervention Method Innovation & BEAMR\cite{landsman_acl22_beamr} (2022), NEUROLOGIC\cite{lu_acl21_neurologic} (2021), NEUROLOGIC AFesque\cite{lu_acl22-neurologic-AFesque} (2022), CD\cite{mudgal_nips23_Controlled-Decoding} (2023), DATG\cite{liang_arxiv24_DATG} (2024) \\ \cline{2-3} 
& Special Issue Resolution & CAT-PAW\cite{gu_acl22-CAT-PAW} (2022), Gemini Discriminator\cite{liu_arxiv22_Gemini} (2022), NADO\cite{meng_NIPS22_NADO} (2022), DECIDER\cite{xu_arxiv24_DECIDER} (2024), ILC\cite{zheng_acl23_ILC} (2023) \\ \hline
\multirow{1}{*}{\centering \makecell{\textbf{CC-LM Guidance}}} & CC-LM Guidance & GeDi\cite{krause_emnlp21_gedi} (2021), DExperts\cite{liu_acl21_DExperts} (2021), MARCO\cite{hallinan_acl23_MARCO} (2023), Air-Decoding\cite{zhong_acl23_Air-Decoding} (2023), Arithmetic\cite{dekoninck_iclr24_Arithmetic} (2024) \\ \hline
\multirow{2.5}{*}{\centering \makecell{\textbf{Model Self-Feedback}}} 
& Inverse Prompting & Inverse Prompting\cite{zou_KDD21_Inverse-Prompting} (2021), Self-Diagnosis and Self-Debiasing (SD)\cite{schick_tacl21_SD} (2021) \\ \cline{2-3} 
& Contrastive Decoding & PREADD\cite{pei_acl23_PREADD} (2023), COGNACGEN\cite{chen_arxiv22_COGNACGEN} (2022), ROSE\cite{zhong_arxiv24_ROSE} (2024) \\ \hline
\multirow{3}{*}{\centering \makecell{\textbf{Energy-Based Model}}} 
& Gradient Sampling & MUCOCO\cite{kumar_nips21_MUCOCO} (2021), MUCOLA\cite{kumar_emnlp22_MUCOLA} (2022), COLD\cite{qin_NEURIPS22_COLD} (2022), COLD-Attack\cite{guo_arxiv24_ColdAttack} (2024), BOLT\cite{liu_acl23_bolt} (2023) \\ \cline{2-3} 
& Acceptance-Rejection Sampling & Mix\&Match\cite{mireshghallah_acl22_mixandmatch} (2022), BlockMH\cite{forristal_conll23_BlockMH} (2023), ScoPE\cite{yu_acl24_ScoPE} (2024) \\ \hline
\multirow{3}{*}{\centering \makecell{\textbf{External Knowledge}}} 
& Semantic Guidance & LM-Steer \cite{han_acl24_LM-Steer} (2024), K2T\cite{pascual_emnlp21_K2T} (2021) \\ \cline{2-3} 
& Knowledge Retrieval & kNN-LM\cite{Khandelwal_iclr20_kNN-LM} (2020), kNN-SCG\cite{trotta_gem22_kNN-SCG} (2022), kNN-CTG\cite{nawezi_tllm23_kNN-CTG} (2023), MEGATRON-CNTRL\cite{xu_emnlp20_MEGATRON-CNTRL} (2020),  GRACE\cite{wen_acl23_GRACE} (2023), Goodtriever\cite{pozzobon_emnlp23_goodtriever} (2023) \\ \hline
\end{tabular}
\end{table}

\subsubsection{\textbf{Classifier Guidance}} 
Classifier Guidance techniques use external classifiers during decoding to introduce control conditions that adjust the output of the language model, enabling control over specific attributes in the generated text. The classifier, broadly defined as a scorer, can be a reward model, neural network, or API.

PPLM (Plug and Play Language Model) \cite{dathathri_iclr20_PPLM} was one of the earliest methods for decoding-time intervention, combining pre-trained language models with attribute classifiers. PPLM controls text attributes, such as topic or sentiment, by adjusting the hidden layer activations using gradients from the attribute classifier. This method guides text generation without modifying the language model, although it may occasionally reduce text fluency. PPLM's flexibility allows it to combine multiple controllers for complex text control.

At each generation step \( t \), PPLM adjusts the direction of historical activations \( H_t \) to control the language model's output:
\begin{equation}
\Delta H_t = \Delta H_t + \alpha \frac{\nabla_{\Delta H_t} \log p(a|H_t + \Delta H_t)}{\|\nabla_{\Delta H_t} \log p(a|H_t + \Delta H_t)\|^\gamma}
\end{equation}
where \( \alpha \) is the step size and \( \gamma \) is the normalization coefficient. After updating \( \Delta H_t \), the language model executes a forward pass to obtain the updated logits \( \tilde{o}_{t+1} \):
\begin{equation}
\tilde{o}_{t+1}, H_{t+1} = \text{LM}(x_t, \tilde{H}_t), \quad \tilde{H}_t = H_t + \Delta H_t
\end{equation}
These logits generate a new probability distribution \( \tilde{p}_{t+1} \), from which the next word is sampled.

FUDGE (Future Discriminators for Generation) \cite{yang_acl21_fudge} offers a simpler and more effective approach than PPLM by dynamically adjusting the probability distribution during generation. FUDGE predicts the attribute probability of the sequence being generated and modifies the logits to align with the expected attribute. Specifically, FUDGE models the text sequence generation as \(P(x_i|x_{1:i-1})\) and adjusts it using Bayesian factorization:
\[
P(x_i|x_{1:i-1}, a) \propto P(a|x_{1:i})P(x_i|x_{1:i-1})
\]
where \(P(a|x_{1:i})\) is modeled by a binary classifier. The output is multiplied with the base model's probabilities to control the attribute during generation. 

\begin{figure}[h]
    \centering
    \includegraphics[width=0.8\textwidth]{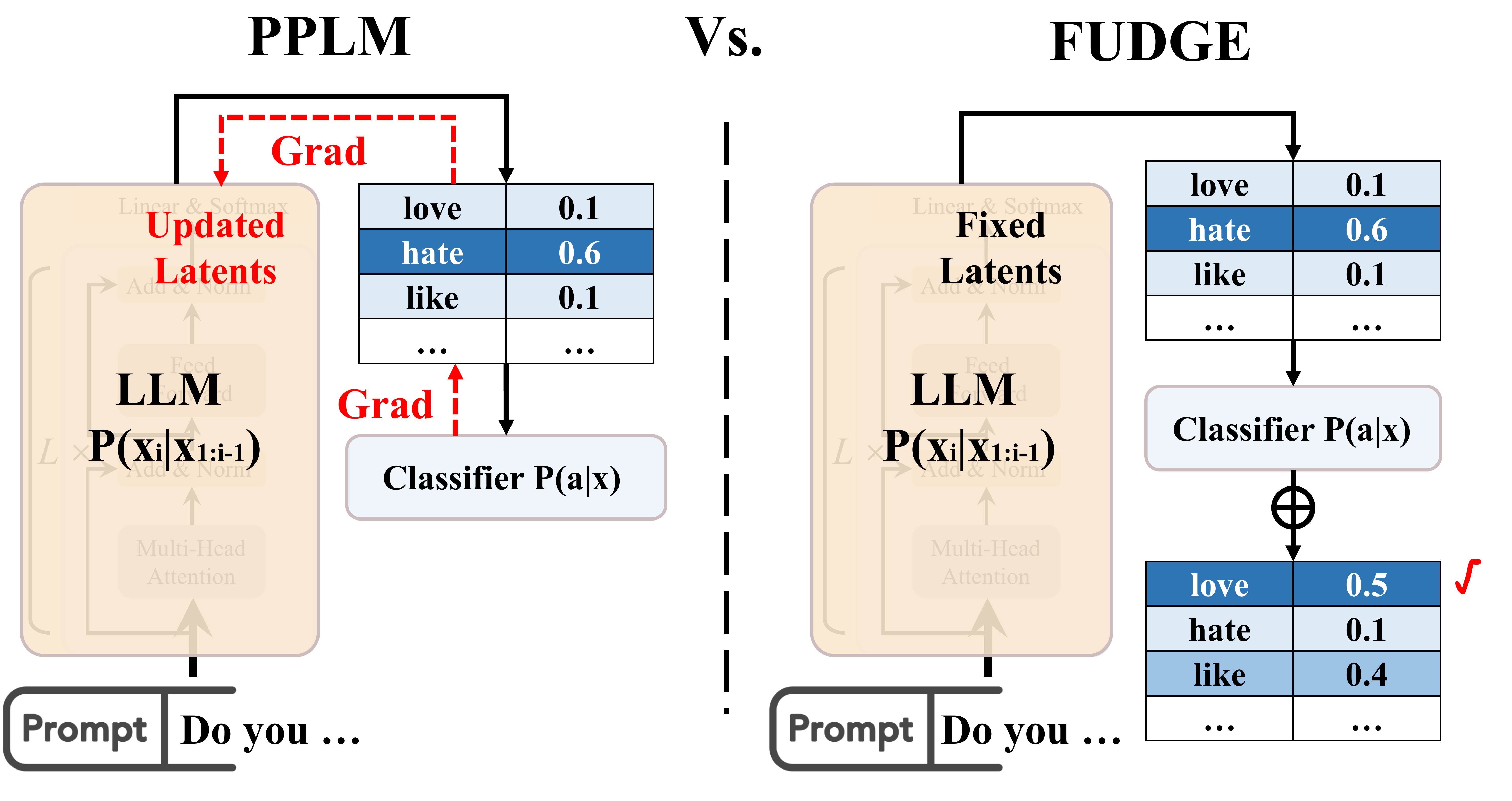}
    \caption{PPLM vs FUDGE}
    \label{fig:pplm_vs_fudge}
\end{figure}

As shown in Figure \ref{fig:pplm_vs_fudge}, FUDGE simplifies the control process compared to PPLM, offering more precise control over text attributes. While both methods use external classifiers for controllable inference, PPLM adjusts hidden states using backpropagation, whereas FUDGE directly modifies logits for attribute control.

CAIF (Classifier-Augmented Inference Framework) \cite{sitdikov_arxiv22_CAIF}, similar to FUDGE, controls text generation by adjusting logits using an external classifier. CAIF offers greater flexibility and adaptability to any existing classifier, effectively controlling specific attributes.

As mentioned earlier, any scorer capable of evaluating a desired attribute can be used for knowledge injection, helping LLMs generate text that meets control conditions. Various scorers have been applied in decoding-time control.
CriticControl \cite{kim_acl23-CriticControl} combines reinforcement learning with weighted decoding, using a critic network to dynamically predict the value of each token based on the generated text's state and reweight probabilities to ensure alignment with desired attributes.
RAD (Reward-Augmented Decoding) \cite{deng_acl23_rad} uses a unidirectional reward model to adjust token probabilities during decoding. It scores each token's contribution to the target attribute and adjusts sampling probabilities for efficient attribute control.
MIL-Decoding (Multiple Instance Learning Decoding) \cite{zhang_acl23_MIL-Decoding} applies multiple instance learning (MIL) to learn toxicity scores at the token level. By combining token toxicity scores with contextual information, it dynamically adjusts the token probability distribution.
SF-GEN (Successor Features Generation) \cite{cao_arxiv23_SF-GEN} separates the language model's dynamics from task-specific rewards using successor features, enabling multi-agent control with a single tensor multiplication, significantly reducing computational overhead.

The aforementioned methods primarily innovate at the scoring model level, often using weighted decoding for knowledge injection. However, other approaches employ diverse decoding techniques to control text generation.
BEAMR (Beam Reweighing) \cite{landsman_acl22_beamr} adjusts beam search candidates by reweighting them based on scores from an attribute classifier, which are used to modify generation probabilities.
NEUROLOGIC \cite{lu_acl21_neurologic} and NEUROLOGIC AFesque \cite{lu_acl22-neurologic-AFesque} use heuristic search to guide text generation under complex lexical constraints.
CD (Controlled Decoding) \cite{mudgal_nips23_Controlled-Decoding} controls text generation with a prefix scoring method. It trains the prefix scorer offline using policy optimization, guiding generation during inference based on the expected reward of partially decoded sequences.
DATG (Dynamic ATtribute Graphs-based CTG) \cite{liang_arxiv24_DATG} employs dynamic attribute graphs to adjust the occurrence of attribute-related keywords, thereby achieving control over text generation.

Several methods have been optimized to address specific challenges in decoding-stage control. For example, CAT-PAW \cite{gu_acl22-CAT-PAW} introduces a lightweight regulator that dynamically adjusts control signals at different decoding positions, mitigating issues of incoherence and repetition when control strength increases.
Gemini \cite{liu_arxiv22_Gemini} uses feature extraction and attribute-driven kernel sampling to address inconsistencies between training and inference features, ensuring the quality of generated text.
NADO (NeurAlly-Decomposed Oracle) \cite{meng_NIPS22_NADO} focuses on complex constraints by decomposing sequence-level constraints into token-level guidance, enabling fine-grained control.
DECIDER \cite{xu_arxiv24_DECIDER} enhances logicality and scientific accuracy by combining language model probability distributions with logical reasoning vectors using first-order logic rules.
ILC (Invariant Learning Characterization) \cite{zheng_acl23_ILC} leverages invariant learning to improve the generalization of attribute predictions across different distributions, ensuring consistency in multi-domain generation.

\subsubsection{\textbf{Class-Conditioned Language Model Guidance}} 
Class-Conditioned Language Models (CC-LMs) use pre-trained or fine-tuned models during decoding to control the attributes of generated text. CC-LMs are trained with specific labels or class information, enabling them to generate text that reflects predefined attributes, such as sentiment or theme. However, directly using these models often yields suboptimal results. To enhance control, the logits from CC-LMs, which contain attribute information, are used as guidance during the decoding process, improving the controlled generation of LLMs.

GeDi (Generative Discriminator) \cite{krause_emnlp21_gedi} is a method that uses class-conditioned language models for text generation control. It fine-tunes a CC-LM using control codes, allowing it to distinguish and generate text with desired attributes.

GeDi applies Bayes' rule during decoding by combining the outputs of a base language model (LM) and a CC-LM to calculate the probability of generating the next token:
\begin{equation}
P(x_t|x_{<t}, c) \propto P_{\text{LM}}(x_t|x_{<t}) P_\theta(c|x_t, x_{<t})^\omega,
\end{equation}
where \(P_{\text{LM}}(x_t|x_{<t})\) is the generation probability from the base LM, and \(P_\theta(c|x_t, x_{<t})\) is the classification probability that the text, after generating \(x_t\), belongs to the control condition \(c\). The parameter \(\omega\) adjusts the bias towards the target attribute.

GeDi enhances control precision by calculating and normalizing the probabilities of the next token under desired and undesired attributes:
\begin{equation}
P_\theta(c|x_{1:t}) = \frac{P(c) \prod_{j=1}^t P_\theta(x_j|x_{<j}, c)}{\sum_{c' \in \{c, \bar{c}\}} P(c') \prod_{j=1}^t P_\theta(x_j|x_{<j}, c')}.
\end{equation}
This guides the base LM's output to align better with the target attribute.

\begin{figure*}[htbp]
    \centering
    \includegraphics[width=\textwidth]{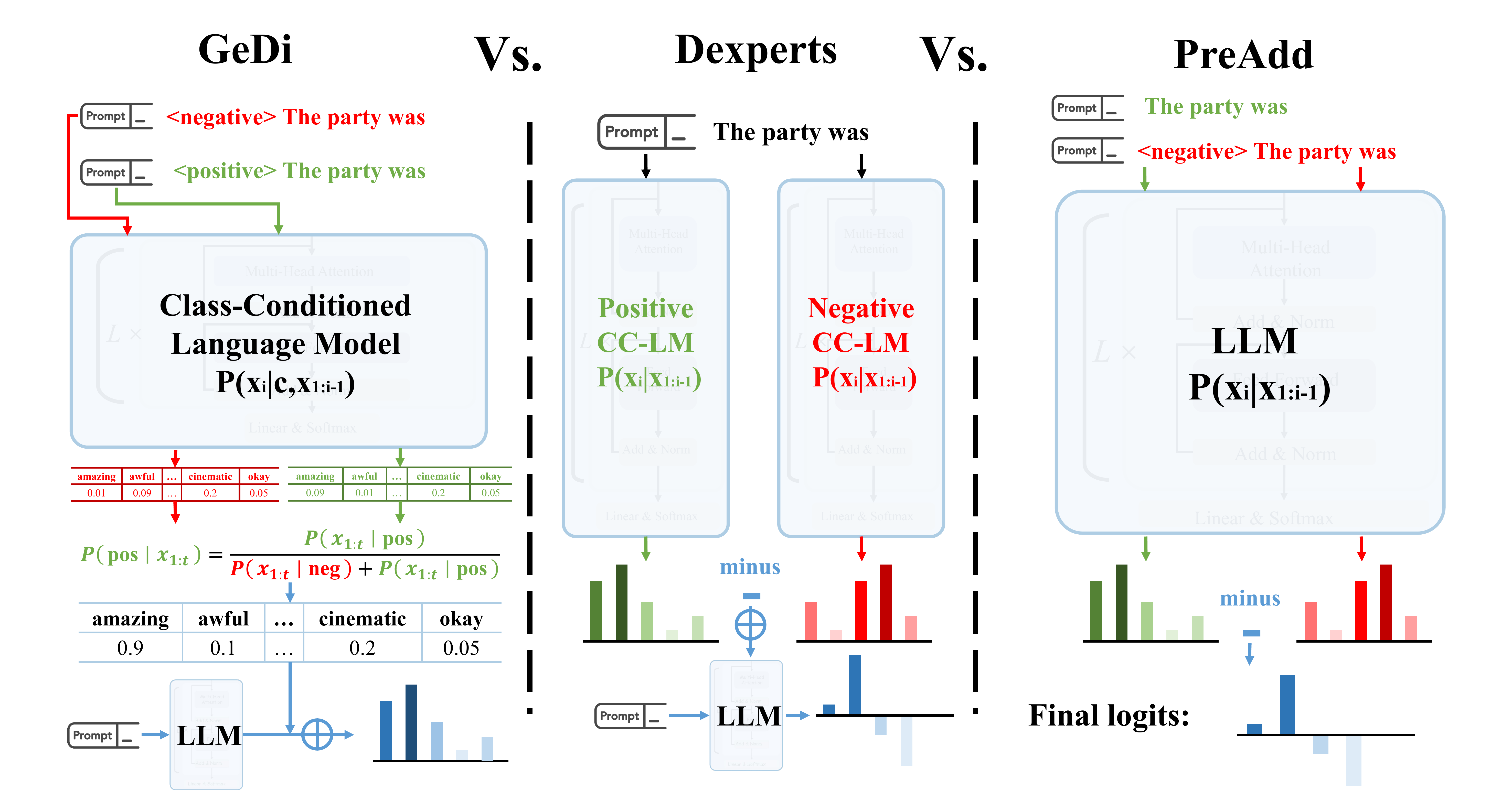}
    \caption{GeDi vs DExperts vs PREADD}
    \label{fig:gedi_vs_dexperts_vs_preadd}
\end{figure*}

DExperts (Decoding-time Experts) \cite{liu_acl21_DExperts} offers a more straightforward contrastive decoding approach by modifying a pre-trained LM's predictions using expert and anti-expert models. DExperts operates on a pre-trained LM \( M \), with an expert model \( M' \) and an anti-expert model \( M'' \), which model text with and without the target attribute, respectively. At time step \( t \), these models produce logits \( z_t \), \( z_t' \), and \( z_t'' \):
\begin{equation}
\tilde{P}(x_t | x_{<t}) = \text{softmax}(z_t + \alpha (z_t' - z_t'')),
\end{equation}
where \(\alpha\) controls the strength of the modification. DExperts adjusts the logits from the base LM using the expert model to align with the target attribute, while the anti-expert model attenuates unwanted attributes. Figure \ref{fig:gedi_vs_dexperts_vs_preadd} illustrates the differences between GeDi, DExperts, and the self-feedback guidance method PREADD (Prefix-Adaptive Decoding) \cite{pei_acl23_PREADD}.

MARCO (Mask and Replace with Context) \cite{hallinan_acl23_MARCO} focuses on correcting text rather than generating it. MARCO trains expert and anti-expert models to detect and replace toxic components during text generation.
Arithmetic \cite{dekoninck_iclr24_Arithmetic} uses model arithmetic techniques for precise attribute control in text generation. It combines multiple models and attributes, including classifiers and class-conditioned language models, through weighted linear combinations and joint operators to optimize and integrate different input distributions.

Air-Decoding \cite{zhong_acl23_Air-Decoding} addresses "attribute collapse," where strong attribute control can impair fluency. Air-Decoding reconstructs the attribute distribution during generation, adjusting token weights using attribute distributions from prefix tuning. This balances attribute-specific and non-attribute words, ensuring the text meets attribute requirements while maintaining fluency.

\subsubsection{\textbf{Self-Feedback Guidance}} 
Self-Feedback Guidance leverages the internal knowledge of pre-trained language models to control and guide text generation \cite{liang_arxiv24_ICSF}. The premise is that while the model has the knowledge to solve a task, it may fail to achieve CTG due to inadequate prompts or output limitations. These methods adjust the generated text during decoding by tapping into the model's inherent knowledge, ensuring alignment with desired attributes.

Inverse Prompting \cite{zou_KDD21_Inverse-Prompting} enhances consistency in text generation by using the generated text to inversely predict the prompt during the generation process. It calculates the conditional probability of the original prompt under the inverse prompt to ensure high consistency between the generated text and the initial prompt.

For example, a traditional model might generate an answer in the format “Question: \${Question} Description: \${Description} Answer: \${Answer}.” In Inverse Prompting, the generated answer is used as a prompt to inversely predict the question, forming an inverse prompt like "\${Answer} answered the question \${Question}." The process involves:

\begin{itemize}
    \item The base language model first generates an answer, e.g., for "What is inverse prompting?", it might generate "Inverse prompting is a method of using generated text to predict prompts."
    \item The answer is then recombined with the question to form the inverse prompt: "Inverse prompting is a method of using generated text to predict prompts. It answered the question 'What is inverse prompting?'"
    \item The conditional probability of the original prompt under the inverse prompt, \(P(c'_p | c'_g)\), where \(c'_p\) is the original prompt and \(c'_g\) is the inverse prompt, is calculated to adjust the scores of generated candidates.
\end{itemize}

Beam search techniques are used during decoding to synthesize candidate scores, allowing the selection of the generated text that best matches the initial prompt, thereby enhancing consistency between the generated content and control attributes.

SD (Self-Diagnosis and Self-Debiasing) \cite{schick_tacl21_SD} leverages the model's ability to self-diagnose and self-debias to identify and reduce biases in generated text. During decoding, SD adjusts word probability distributions to minimize biased content. The self-diagnosis process in SD is conceptually similar to Inverse Prompting, and its self-debiasing approach was one of the earliest applications of contrastive decoding for detoxification control.

Contrastive Decoding Approaches play a significant role in self-feedback guidance by comparing the logits generated for different prompts during decoding, enabling flexible control over text generation attributes. These methods often involve designing prompts that induce the model to generate text with attributes opposite to those desired, using this comparison to guide the generation of text that aligns with the intended attributes.

PREADD (Prefix-Adaptive Decoding) \cite{pei_acl23_PREADD} controls text generation attributes by comparing and adjusting the logits generated by different prompts. During the generation process of model G, PREADD pre-adds a prefix \( r_{1:k} \) and adjusts the output by comparing the logit difference \( d \) between the prefixed and non-prefixed outputs:
\begin{equation}
d := \log P(x_{i+1} | r_{1:k}, x_{1:i}) - \log P(x_{i+1} | x_{1:i})
\end{equation}
This difference \( d \) is applied with a multiplier \( \alpha \) to control the output intensity, allowing the model to adjust attribute control flexibly. The final probability model is:
\begin{equation}
P(x_{i+1} | r_{1:k}, x_{1:i})^\alpha P(x_{i+1} | x_{1:i})^{1-\alpha}
\end{equation}

For example, in detoxification tasks, PREADD uses a static prefix \( e_{1:m} \) that encourages the generation of toxic text, such as: "The following text perpetuates negative stereotypes, is threatening or sexually explicit, or contains profane language." By calculating the logit differences between the prefixed and non-prefixed prompts at each generation step, PREADD effectively adjusts the attributes of the generated text.

COGNACGEN \cite{chen_arxiv22_COGNACGEN} and ROSE (Reverse Prompt Contrastive Decoding) \cite{zhong_arxiv24_ROSE} share similar ideas with SD and PREADD. COGNACGEN adjusts token generation by generating guiding words that align with complex constraints, integrating this guidance through prefix adjustment. ROSE uses reverse prompts to induce harmful responses, applying them during inference to suppress undesirable content, enhancing output safety.

As discussed earlier, spurious correlations occur when models mistakenly identify unrelated features as important, leading to biased attribute selection in text generation. This issue also affects CTG during decoding.
SCM (Structural Causal Model) \cite{zhiting_nips21_SCM} reduces bias by incorporating causal reasoning into text generation, allowing for attribute modification while preserving other features through counterfactual inference. FPT (Focused Prefix Tuning) \cite{ma_acl23_FPT} addresses interference from implicit attributes by using specific and general prefixes, training them separately and combining their logits to enhance control over explicit attributes.

\subsubsection{\textbf{Energy-Based Model Guidance}}
Energy-Based Model (EBM) Guidance methods control the attributes of generated text by optimizing an energy function during the generation process. These methods assign lower energy values when specific constraints are met, guiding the text to align with desired attributes. EBMs are often used to balance multiple attributes, searching for decoding strategies that satisfy these constraints within the energy space.

EBM-guided generation relies on the sampling method. When sampling from the joint distribution of multiple control attributes, the key is to select an optimal sampling method that efficiently identifies the best token within the energy model space. Some methods use gradient information from the energy model to achieve text constraint control by sampling in the solution space.

MUCOCO (Multi-Constraint Controlled Optimization) \cite{kumar_nips21_MUCOCO} was one of the earliest energy-based CTG methods, treating decoding as a continuous optimization problem with multiple differentiable constraints. It combines gradient descent and Lagrange multipliers for multi-attribute control. MUCOLA (Multiple Constraints using Langevin Dynamics) \cite{kumar_emnlp22_MUCOLA} improves upon MUCOCO by integrating the language model's log-likelihood with user-defined constraints into an energy function, using Langevin dynamics for non-autoregressive sampling. COLD (Constrained Decoding with Langevin Dynamics) \cite{qin_NEURIPS22_COLD} also employs Langevin dynamics, iteratively updating to generate text that meets specific constraints. COLD-Attack \cite{guo_arxiv24_ColdAttack} extends COLD by generating adversarial prompts through energy-constrained decoding. To improve sampling efficiency, BOLT (Bias-Optimized Logit Tuning) \cite{liu_acl23_bolt} adds adjustable biases to predicted logits at each decoding step, optimizing them via gradient descent to minimize overall energy and ensure compliance with specified constraints.

Another class of EBM sampling methods uses acceptance-rejection mechanisms, such as Metropolis-Hastings and Gibbs sampling, to control text attributes without relying on gradient information, allowing the use of black-box scorers.

Mix\&Match \cite{mireshghallah_acl22_mixandmatch} combines scores from pre-trained black-box models (e.g., fluency, control attributes, context fidelity) into a unified energy function and uses Metropolis-Hastings sampling to generate text that meets desired attributes. During generation, Mix\&Match incrementally proposes token replacements, accepting changes that lower the energy. BlockMH (Block Metropolis-Hastings Sampler) \cite{forristal_conll23_BlockMH} enhances efficiency and output quality by introducing a block-level proposal sampler that iteratively rewrites the sequence. ScoPE (Score-based Progressive Editor) \cite{yu_acl24_ScoPE} integrates the energy model with the editing process, progressively editing intermediate tokens to align with target attributes and guiding the black-box model to generate the desired text.

\subsubsection{\textbf{External Knowledge Guidance}}
External Knowledge Guidance enhances text generation by integrating information from external knowledge bases or retrieval mechanisms. These methods introduce relevant knowledge dynamically, improving coherence and alignment with desired attributes. They can be categorized into two types: semantic guidance and knowledge retrieval.

Semantic Guidance methods incorporate external semantic information and context-relevant information to modulate the model's output. 

K2T (Keyword to Text) \cite{pascual_emnlp21_K2T} ensures the inclusion of specific keywords by adjusting log probabilities based on cosine similarity between words and keywords at each generation step.
LM-Steer \cite{han_acl24_LM-Steer} enables flexible and interpretable control over language model generation styles by applying a learnable linear transformation to output word embeddings.

Knowledge retrieval methods enhance coherence, accuracy, and control by retrieving relevant information from external sources during generation.
kNN-LM \cite{Khandelwal_iclr20_kNN-LM} is an early retrieval-augmented method, building a key-value store from training data and retrieving the k nearest neighbors using context embeddings, interpolating this information into predictions. 
kNN-SCG \cite{trotta_gem22_kNN-SCG} and kNN-CTG \cite{nawezi_tllm23_kNN-CTG} extend kNN-LM by combining retrieval techniques with CTG, enhancing control through relevant example retrieval.
Another notable method, MEGATRON-CNTRL \cite{xu_emnlp20_MEGATRON-CNTRL}, enhances story generation by dynamically integrating keywords and retrieving the relevant knowledge.
GRACE \cite{wen_acl23_GRACE} combines generative and contrastive learning to adjust the relevance and diversity of retrieved content.
Goodtriever \cite{pozzobon_emnlp23_goodtriever} integrates toxic and non-toxic data stores, combining store output with model logits for adaptive toxicity mitigation.

While Decoding-time Intervention offers significant flexibility and allows for real-time adjustments during the text generation process, it typically relies on external models or components to inject the desired control conditions. This dependency can increase inference time due to the additional computation needed to adjust the output. Moreover, directly manipulating the model's output probabilities may disrupt the natural fluency and coherence of the generated text, as these adjustments might force the model to select less likely tokens that fit the control conditions, potentially impacting the text's smoothness.

\subsection{Summary}

Inference-stage methods provide precise control in CTG by dynamically adjusting the generation process. These methods include Prompt Engineering, Latent Space Manipulation, Decoding-time Intervention, and various guidance techniques, each offering distinct advantages and challenges.

\textbf{Prompt Engineering} methods exert control directly at the input level through hard prompts \cite{shin_emnlp20_Autoprompt,ramirez_sigdial23_DAs,zhang_SEM23_PCFG} and soft prompts \cite{li_acl21_PrefixTuning,lester_emnlp21_PromptTuning,liu_arxiv21_PTuning}, without requiring additional model training, making them suitable for quickly adjusting generation strategies. Hard prompts rely on explicit natural language instructions, while soft prompts use trainable vectors for more granular control. Although flexible and resource-efficient, the effectiveness of this approach depends on the model's sensitivity to and accuracy in interpreting the prompts.

\textbf{Latent Space Manipulation} involves introducing control vectors into the model's latent space to adjust the characteristics of the generated text \cite{chan_nips21_GENhance,subramani_acl22_LatentStreeringVectors,liu_arxiv24_ICV,turner_arxiv24_actadd,konen_acl24_StyleVectors}. By directly manipulating the model's activation states, this method allows for precise control, especially in multi-attribute tasks.

\textbf{Decoding-time Intervention} uses dynamic adjustments during the decoding process to control the generated output, including classifier guidance \cite{dathathri_iclr20_PPLM,yang_acl21_fudge,sitdikov_arxiv22_CAIF}, class-conditioned language models \cite{krause_emnlp21_gedi,liu_acl21_DExperts,hallinan_acl23_MARCO}, energy-based models \cite{kumar_nips21_MUCOCO,kumar_emnlp22_MUCOLA,mireshghallah_acl22_mixandmatch}, model self-feedback \cite{schick_tacl21_SD,zhong_arxiv24_ROSE}, and external knowledge \cite{nawezi_tllm23_kNN-CTG,pozzobon_emnlp23_goodtriever}. Adjusting output probabilities during generation enables complex attribute control but may impact text naturalness and coherence, and adds computational complexity due to reliance on external models.

Overall, inference-stage methods provide flexible and dynamic text control capabilities, enabling highly customized text generation without altering the original model structure. However, they often rely on external resources and models, which may pose challenges in terms of fluency and consistency. Nevertheless, these methods excel in scenarios requiring attribute control.

\section{Evaluation}
\label{sec:ctg_eval}

Evaluation metrics for CTG tasks can be broadly categorized into three types: automatic evaluation, human evaluation, and LLM-based evaluation methods, as shown in Table \ref{tab:evaluation}.
\begin{table}[htbp]
\centering
\caption{Summary of Evaluation Methods and Metrics}
\label{tab:evaluation}
\renewcommand{\arraystretch}{1.6}
\footnotesize
\begin{tabular}{llp{0.55\textwidth}}
\hline
\textbf{Evaluation Type}             & \textbf{Aspect}                & \textbf{Description}                                   \\ \hline
\multirow{7}{*}{\makecell[l]{\textbf{Automatic} \\ \textbf{Evaluation}}}  & \multirow{5}{*}{\makecell[l]{\textbf{General Metrics}}} & N-gram Overlap-based: BLEU\cite{papineni_acl02_bleu}, ROUGE\cite{lin_acl04_rouge}, METEOR\cite{banerjee_acl05_meteor}, NIST\cite{doddington_hlt02_nist}, Distinct-n\cite{li_naacl16_diversity}, Repetition-n\cite{shao_emnlp19_repetition}, Self-BLEU\cite{zhu_sigir18_selfbleu}               \\ \cline{3-3}
                                     &                                 & Language Model-based: Perplexity, BertScore\cite{zhang_iclr20_bertscore}, MoverScore\cite{zhao_emnlp19_moverscore}, BLEURT\cite{sellam_acl20_bleurt} \\ \cline{3-3}
                                     &                                 & Distance-based: TER\cite{snover_amta06_ter}                                           \\ \cline{3-3}
                                     &                                 & Other: CIDEr\cite{vedantam_cvpr15_cider}, SPICE\cite{anderson_eccv16_spice}      \\ \cline{2-3}
                                     & \makecell[l]{\textbf{Task-specific Metrics}}          & Classifiers or API for specific attributes\cite{zhong_acl23_Air-Decoding, liang_arxiv24_DATG}                    \\ \hline
\multirow{2.5}{*}{\makecell[l]{\textbf{Human} \\ \textbf{Evaluation}}}      & \multirow{1.5}{*}{\makecell[l]{\textbf{Evaluation Metrics}}}               & Fluency, Coherence, Topicality, General Quality, Attribute Relevance \\ \cline{2-3}
                                     & \textbf{Evaluation Methods}               & A/B test, N-point Likert-like scale                           \\ \hline
\textbf{LLM-based}                   & \textbf{Approach}              & Using LLM for Evaluation\cite{liu_arxiv24_ICV, dai_iclr24_SafeRLHF, li_arxiv24_DESTEIN, zhong_arxiv24_ROSE, wang_arxiv24_InferAligner, guo_arxiv24_ColdAttack, xia_arxiv24_fofo}                                      \\ \hline
\end{tabular}
\end{table}

\subsection{Automatic Evaluation}
Automatic evaluation uses specific metrics or models and can be divided into general and task-specific evaluations. General metrics assess overall text quality across various CTG tasks, while task-specific evaluations focus on quality based on specific attributes.

\subsubsection{\textbf{General Metrics}}
Depending on how they are calculated, general metrics can be divided into n-gram overlap-based metrics, language model-based metrics, distance-based metrics, etc.

\textbf{N-gram Overlap-Based Metrics:} These metrics convert text into sets of n-gram units and focus on the similarity of n-gram distributions, typically by comparing generated text to reference text.

\textbf{BLEU\cite{papineni_acl02_bleu}:} BLEU is a common evaluation metric that measures the similarity between generated text and reference text, focusing on precision. It calculates the proportion of n-gram units in the generated text that appear in the reference text, with the formula as follows:

\begin{equation}
\text{BLEU-n} = \frac{
    \sum_{c \in C} \sum_{g \in c} \text{Count}_{\text{clip}}(g)
}{
    \sum_{c' \in C} \sum_{g' \in c'} \text{Count}(g')
}
\end{equation}
where \( C \) is the set of candidate texts and \( g \) is an n-gram. \(\text{Count}_{\text{clip}}(g)\) is the n-gram's count in the reference text, capped by its count in the candidate. \(\text{Count}(g')\) is the total n-gram count in the candidate. A higher value indicates greater similarity between the generated and reference texts.

\textbf{ROUGE\cite{lin_acl04_rouge}:} ROUGE is conceptually similar to BLEU but calculates the proportion of n-grams in the reference text that appear in the generated text, focusing on recall rather than precision.

\begin{equation}
\text{ROUGE-n} = \frac{
    \sum_{r \in R} \sum_{g \in r} \text{Count}_{\text{match}}(g)
}{
    \sum{g \in r} \text{Count}(g)
}
\end{equation}
where \( R \) denotes the set of reference texts, \( r \) represents a reference text, and \( g \) denotes an n-gram. \(\text{Count}_{\text{match}}(g)\) represents the number of matching n-grams in the generated text, and \(\text{Count}(g)\) represents the total count of n-grams in the reference text. The higher this value, the greater the similarity between the generated and reference texts.

\textbf{METEOR\cite{banerjee_acl05_meteor}:} BLEU focuses on precision, and ROUGE on recall, but both have limitations. METEOR addresses this by combining them into an "F1 score" with the following formula:

\begin{equation}
F_{mean} = \frac{10PR}{R+9P}
\end{equation}
where \( P \) represents precision, and \( R \) represents recall.

Unlike BLEU, which considers only exact n-gram matches, METEOR incorporates additional mechanisms like synonym matching and stemming, using resources like WordNet. For example, "journey" and "tour" would be matched as synonyms, improving evaluation accuracy.

Additionally, METEOR considers n-gram alignment between generated and reference texts. It introduces the concept of "chunks," which are continuous sequences of matched n-grams. A penalty is applied for discontinuities in the matching sequences:

\begin{equation}
Penalty = 0.5 \left( \frac{\text{chunks}}{\text{unigrams matched}} \right)^3
\end{equation}
where \(\text{chunks}\) represents the number of discontinuous matched sequences, and \(\text{unigrams matched}\) represents the number of matched words. This penalty reduces the score for excessive discontinuities. The final score is calculated as follows:

\begin{equation}
Score = F_{mean}(1 - Penalty)
\end{equation}
where \( Score \) represents the final METEOR score. More chunks result in a higher penalty and a lower METEOR score. This method better accounts for word order and coherence, offering a more detailed and accurate evaluation than n-gram-based metrics.

\textbf{NIST\cite{doddington_hlt02_nist}:} NIST builds on BLEU by introducing the concept of information weight:

\begin{equation}
\text{Info}(w_1 \ldots w_n) = \log_2 \left( \frac{\text{Count}(w_1 \ldots w_{n-1})}{\text{Count}(w_1 \ldots w_n)} \right)
\end{equation}
where \(\text{Count}(w_1 \ldots w_{n-1})\) represents the occurrence count of the first \(n-1\) words, and \(\text{Count}(w_1 \ldots w_n)\) represents the occurrence count of the full n-gram. Rare n-grams are given higher weight.

NIST assigns varying weights to each n-gram, averaging them for a final score that better evaluates similarity by accounting for rare n-grams.

\textbf{Distinct-n\cite{li_naacl16_diversity}:} Distinct-n measures the diversity of generated text by calculating the ratio of unique n-grams to total n-grams:

\begin{equation}
\text{Distinct-n} = \frac{\text{Count}\left(\text{unique n-gram}\right)}{\text{Count}\left(\text{n-gram}\right)}
\end{equation}
where \(\text{Count}(\text{unique n-gram})\) represents the number of unique n-grams in the generated text, and \(\text{Count}(\text{n-gram})\) represents the total number of n-grams.

\textbf{Repetition-n\cite{shao_emnlp19_repetition}:} Repetition-n indirectly evaluates the diversity of generated text by calculating the ratio of n-grams that occur more than once to the total number of n-grams:

\begin{equation}
\text{Repetition-n} = \frac{\text{Count}\left(\text{repeated n-gram}\right)}{\text{Count}\left(\text{n-gram}\right)}
\end{equation}
where \(\text{Count}(\text{repeated n-gram})\) represents the number of repeated n-grams in the generated text, and \(\text{Count}\left(\text{n-gram}\right)\) represents the total number of n-grams. This ratio assesses the repetition level of the generated text, reflecting its diversity.

\textbf{Self-BLEU\cite{zhu_sigir18_selfbleu}:} Self-BLEU measures diversity by calculating BLEU scores between generated texts, not against references. It averages BLEU scores across generated texts and lower Self-BLEU scores indicate higher diversity among the generated texts.

\textbf{Language Model-Based Metrics:} 

\textbf{Perplexity\cite{jozefowicz_arxiv16_perplexity}:} Perplexity measures the model's ability to predict test data, indicating the model's uncertainty in its predictions. In NLP tasks, perplexity represents the model's accuracy in predicting word sequences in a test set. It is calculated as follows:

\begin{equation}
\text{PPL} = \left( \prod_{i=1}^{n} \frac{1}{p(w_i | w_1, w_2, \ldots, w_{i-1})} \right)^{\frac{1}{n}}
\end{equation}

In practice, a proxy model (e.g., GPT-2) is often used to calculate the perplexity of the generated text. Lower PPL indicates higher fluency of the generated text.

\textbf{BertScore\cite{zhang_iclr20_bertscore}:} BertScore is a language generation evaluation metric based on pre-trained BERT contextual embeddings It computes the similarity of two sentences as a sum of cosine similarities between their tokens’ embeddings. Unlike n-gram-based metrics, BertScore captures semantic information, offering a more accurate evaluation.

\textbf{MoverScore\cite{zhao_emnlp19_moverscore}:} MoverScore combines word embeddings with Earth Mover's Distance (EMD). Unlike BertScore, which considers each word's independent similarity, MoverScore treats text as a distribution of word embeddings and calculates the distance between these distributions, capturing contextual information and word relationships for more accurate evaluation.

\textbf{BLEURT\cite{sellam_acl20_bleurt}:} BLEURT improves upon BertScore by training BERT on synthetic data generated by adding random perturbations to Wikipedia sentences. This allows the metric to be more robust to domain and quality drift, providing higher evaluation accuracy.

\textbf{Distance-Based Metrics:} 

\textbf{TER\cite{snover_amta06_ter}:} TER evaluates the quality of generated text by comparing it with reference text, calculating the number of edit operations (insertion, deletion, substitution, and shift of words) needed to transform the generated text into the reference text. The formula is:

\begin{equation}
\text{TER} = \frac{\text{Number of Edits}}{\text{Average Number of Reference Words}}
\end{equation}
Lower TER indicates higher similarity and quality of the generated text.

\textbf{Other Metrics:} 

\textbf{CIDEr\cite{vedantam_cvpr15_cider}:} CIDEr evaluates the quality of generated text by comparing it with multiple reference texts, incorporating TF-IDF weighting to assign different weights to different n-grams. This highlights important n-grams and reduces the influence of common ones, capturing key content and important information for a more nuanced evaluation.

\textbf{SPICE\cite{anderson_eccv16_spice}:} SPICE is a semantic similarity metric that uses a probabilistic context-free grammar (PCFG) dependency parser to parse generated and reference texts into syntactic dependency trees. These are then mapped to scene graphs, including entities, attributes, and relations, and the similarity score is calculated based on the matching between the scene graphs. Compared to n-gram-based metrics, SPICE better captures semantic information.

\subsubsection{\textbf{Task-specific Metrics}}
To evaluate whether the generated text meets the specified attributes in CTG tasks, a classifier is often used. This classifier can be obtained by training a base model (e.g., BERT) on a specific dataset (e.g., IMDB). Table \ref{tab:classifier} lists commonly used datasets and base models. Alternatively, existing models can be directly used, often sourced from HuggingFace, such as DistilBERT-base-uncased-finetuned-SST-2 for emotion tasks\footnote{\href{https://huggingface.co/distilbert/distilbert-base-uncased-finetuned-sst-2-english}{https://huggingface.co/distilbert/distilbert-base-uncased-finetuned-sst-2-english}}, tweet-topic-21-multi for topic tasks\footnote{\href{https://huggingface.co/cardiffnlp/tweet-topic-21-multi}{https://huggingface.co/cardiffnlp/tweet-topic-21-multi}}, and the Perspective API for toxicity tasks\footnote{\href{https://perspectiveapi.com/}{Perspective API}}.

\begin{table*}[htbp]
\centering
\caption{Common Base Models and Datasets for Training Classifiers}
\label{tab:classifier}
\renewcommand{\arraystretch}{1.6}
\footnotesize
\begin{tabular}{lp{0.36\textwidth}p{0.48\textwidth}}
\hline
\multicolumn{1}{c}{\textbf{Attribute}} & \multicolumn{1}{l}{\textbf{Base Model}} & \multicolumn{1}{l}{\textbf{Dataset}} \\ \hline

\multirow{1.5}{*}{\makecell[l]{\textbf{Emotion}}} & BERT\cite{devlin_naacl19_bert}, RoBERTa\cite{liu_arxiv19_roberta}, DeBERTa\cite{he_iclr21_deberta}, distilBERT\cite{sanh_arxiv19_distilbert}, MacBERT\cite{cui_emnlp20_revisiting} & IMDB\cite{maas_acl11_learning}, AMAZON-5\cite{mcauley_recsys13_hidden}, SST-5\cite{socher_emnlp13_recursive}, SST-2\cite{socher_emnlp13_recursive}, Yelp\cite{zhang_nips15_character}, Twitter sentiment\cite{barbieri_emnlp20_tweeteval}, DailyDialog\cite{li_ijcnlp17_dailydialog} \\ \hline
\textbf{Topic} & BERT\cite{devlin_naacl19_bert}, RoBERTa\cite{liu_arxiv19_roberta} & AG-NEWS\cite{zhang_nips15_character}, DBpedia\cite{zhang_nips15_character} \\ \hline
\multirow{1.5}{*}{\makecell[l]{\textbf{Toxicity}}} & RoBERTa\cite{liu_arxiv19_roberta}, DeBERTa\cite{he_iclr21_deberta} & Jigsaw Toxic Comment Classification Challenge\cite{cjadams_kaggle18_jigsaw}, RealToxicityPrompts\cite{gehman_emnlp20_realtoxicityprompts} \\ \hline

\end{tabular}
\end{table*}

\subsection{Human Evaluation}
While automated evaluation meets most evaluation requirements, considering the diversity of CTG tasks and the limitations of automated evaluation, human evaluation can serve as a valuable supplement, providing customized assessment and more accurate results. This section introduces the metrics and methods used in human evaluation.

\subsubsection{\textbf{Metric}}
Common human evaluation metrics include:

\textbf{Fluency:} Fluency measures whether the generated text is grammatically correct, easy to understand, and free from repetition.

\textbf{Coherence:} Assesses whether the text maintains a linguistic style, exhibits causal and temporal dependency between sentences, and whether the information is logically organized.

\textbf{Topicality:} Measures consistency with the context of the given prompt.

\textbf{General quality:} Unlike the more holistic metrics mentioned above, this class of metrics is more specific, evaluating particular aspects of the generated text, such as commonsense, logical consistency, diversity of expression, lexical richness, and grammatical correctness.

\textbf{Attribute relevance:} Similar to the metrics in automated evaluation, this metric judges whether the generated text meets the given attribute (e.g., emotion, topic, lexical features).

\subsubsection{\textbf{Method}}
Common human evaluation methods include A/B testing and Likert scales.

\textbf{A/B test:} A/B testing is a comparison-based evaluation method where human annotators are asked to select the text that better meets the requirements from two (or more) generated texts based on a given question (e.g., which sentence is more logical?).

\textbf{N-point Likert-like scale:} The N-point Likert-like scale is a quantitative evaluation method where human annotators rate the generated text according to predefined scoring standards (usually discrete), such as 0 representing low quality and 3 representing high quality.

\subsection{LLM-based Evaluation}
With the advent of powerful language models like ChatGPT, LLM-based evaluation methods are becoming increasingly popular \cite{liu_arxiv24_ICV,dai_iclr24_SafeRLHF,li_arxiv24_DESTEIN,zhong_arxiv24_ROSE,wang_arxiv24_InferAligner,guo_arxiv24_ColdAttack,xia_arxiv24_fofo}. These evaluation methods only require the construction of specific prompts, allowing the model to evaluate the generated text. Compared to traditional automated evaluation methods, LLM-based methods are more diverse, meeting specific evaluation needs and returning richer evaluation results. Compared to human evaluation methods, LLM-based methods are more practical, significantly reducing evaluation costs (e.g., labor, time, money) while also reducing the impact of human annotators' subjective biases to some extent.

\subsection{Benchmarks}

Several benchmarks have been proposed in the research of CTG evaluation to assess the performance of generation models under different tasks and conditions.

\begin{itemize}
    \item \textbf{CTRLEval} \cite{ke_acl22_CTRLEval} introduces an unsupervised, reference-free metric to evaluate controlled text generation quality, using text infilling with a pre-trained model (e.g., PEGASUS) to assess coherence, consistency, and attribute relevance.

    \item \textbf{ConGenBench} \cite{ashok_arxiv24_ConGenBench} benchmarks controllable generation methods by generating constrained datasets with instruction-tuned LLMs, showcasing their potential, particularly in style tasks.

    \item \textbf{CoDI-Eval} \cite{chen_AAAI24_CoDIEval} integrates diverse instructions by expanding human-written seeds, introducing new tasks and standards for testing LLMs' controllable generation in complex settings.

    \item \textbf{FOFO} \cite{xia_arxiv24_fofo} is a benchmark developed through AI-human collaboration, covering a variety of real-world formats and instructions to evaluate LLMs' format adherence capabilities.
\end{itemize}

\section{Applications}
\label{sec:ctg_app}

CTG technology has developed diverse control generation methods to meet various generation needs across different fields. These methods can be categorized into vertical domain applications and general task applications. Vertical domain applications are tailored to specific tasks within particular industries, focusing on specialization and precision, while general task applications address cross-domain needs, offering high versatility. The following sections provide an overview and analysis of CTG technology in different application scenarios.

\subsection{Vertical Domain Applications}
CTG has shown strong adaptability in specialized fields, effectively addressing unique generation needs in domains such as news reporting, scientific literature, and educational content creation. By employing specialized models and methods, CTG enhances the quality and relevance of generated text, making it more targeted and professional.

In news generation, DeepPress\cite{rahali_2023_DeepPress} integrates pre-trained models to produce topic-aware news content, enhancing objectivity and coherence, while SeqCTG\cite{spangher_acl22_SeqCTG} ensures logical consistency in articles using local control codes. For scientific texts, MReD\cite{shen_acl22_MReD} utilizes structured datasets to improve the domain specificity of generated content.

In education, CE (Complexity Embedding)\cite{jinran_ccl23_CE} leverages complexity embeddings to control lexical complexity, enabling the creation of customized learning materials for language learners. For multilingual generation, SweCTRL-Mini\cite{kalpakchi_arxiv23_SweCTRL-Mini} applies control codes in Swedish text generation, while Collocation2Text\cite{Vychegzhanin_arxiv22_Collocation2Text} guides Russian text generation through specified phrases.

CTG also enhances internet text generation. PCTG-X\cite{yang_2024_PCTGX} uses text prompts and attribute labels to control the stance and style of social media content, while CounterGeDi\cite{saha_ijcai22_CounterGeDi} suppresses unwanted attributes to counter hate speech. In Chinese content, CAT-LLM\cite{tao_arxiv24_CAT} facilitates style transformation using LLMs and text style modules.

In niche applications like recipe generation, RecipeWithPlans\cite{liu_acl22_RecipeWithPlans} combines content planning with sequence generation to produce coherent and logically structured recipes.

\subsection{General Task Applications}
General task applications address cross-domain challenges like toxicity removal, dialogue generation, and story creation, making these methods applicable across various scenarios.

In toxicity control, SRDT\cite{leong_acl23_SRDT} manipulates attention layers to reduce toxic content, while DESTEIN\cite{li_arxiv24_DESTEIN} and InferAligner\cite{wang_arxiv24_InferAligner} adjust activation states to lower the likelihood of generating harmful content.
Additionally, UncertaintyAttack\cite{zeng_arxiv24_UncertaintyAttack} exploits changes in the probability distribution of model output logits to carry out security attacks, highlighting the threat that improper application of CTG poses to the reliability of LLMs.

For dialogue generation, Personalized-Dialogue\cite{zheng_aaai20_Personalized-Dialogue} enhances personalization by incorporating user data, and MultiT-C-Dialog\cite{zeng_-naacl21_MultiT-C-Dialog} employs multi-task learning to improve dialogue quality. ECCRG\cite{chen_2024_ECCRG} enhances emotional expression and coherence through emotion and content control.

In story generation, Plug-and-Blend\cite{lin_nuse21_PlugandBlend} offers fine control over multiple themes, while CHAE\cite{wang_coling22_CHAE} allows detailed customization of characters and emotions. SCSC\cite{cho_www22_SCSC} ensures consistency and diversity in storytelling, and PMCSG\cite{Vychegzhanin_2024_PMCSG} generates narratives that meet key plot points by selecting paths with minimal perplexity.

In keyword-controlled generation, Keyword Position\cite{sasazawa_siggen23_KeywordPosition} enhances alignment with user intent by controlling keyword placement, making it suitable for tasks like automated summary generation.

\section{Challenges and Appeals}
\label{sec:ctg_c&a}

\subsection{Challenges}

\subsubsection{\textbf{Reduced Fluency and Practicality}}
Despite the remarkable progress in LLMs like GPT-3 and BERT, challenges remain in achieving fluency and practicality in generated text. Issues such as incoherence, semantic ambiguity, or redundancy often arise, particularly in complex tasks or when precise responses are required. These shortcomings can significantly diminish the practical value of the generated content \cite{zhong_acl23_Air-Decoding,liang_arxiv24_DATG}. Therefore, enhancing the fluency and practical application of generated text remains a critical challenge.

\subsubsection{\textbf{Complexity of Multi-Attribute Control}}
Controlling multiple attributes simultaneously, such as emotion, style, and topic, poses a significant challenge due to the complex interdependencies and constraints among these attributes. While current research mainly focuses on single-attribute control, multi-attribute control is still in its early stages \cite{gu_emnlp22_Discrete}. The ability to precisely control multiple attributes while maintaining the quality of generated text is an unresolved issue that would greatly enhance the customization and utility of AI-generated content.

\subsubsection{\textbf{Incomplete Attribute Decoupling}}
Attribute decoupling, the ability to control one attribute without affecting others, remains an ongoing challenge due to the presence of spurious correlations. Current methods struggle to achieve complete attribute decoupling in practice \cite{zhiting_nips21_SCM}. For example, altering the sentiment of a text might inadvertently shift its focus to a particular topic, such as politics. Achieving complete decoupling to ensure the independence and stability of multi-attribute control is a key research direction.

\subsubsection{\textbf{Decoding Time Optimization}}
Decoding time, or the time required for a model to generate text, is a crucial performance indicator for the practical application of AI-generated content. The large parameter sizes of current LLMs often result in a time-consuming generation process, affecting their feasibility in real-time applications. This issue is particularly relevant when generating long texts or requiring multiple iterations. Thus, significantly reducing decoding time without compromising text quality is a major challenge that necessitates in-depth research into model architecture optimization and improvements in decoding algorithms.

\subsubsection{\textbf{Lack of Precision in Content Control}}
Achieving precise content control, or hard control, in CTG remains challenging. While existing models can generate text that meets expectations to some extent, they often fall short in accuracy. For instance, in tasks requiring strict lexical control, model performance is often unsatisfactory \cite{ashok_arxiv24_ConGenBench}.

\subsection{Appeals}

\subsubsection{\textbf{Research Should Be More Oriented Towards Real-World Applications}}
Many decoding-phase methods face limitations in practicality, particularly in balancing time efficiency with effectiveness. Future research should prioritize practical application needs, aiming to strike an optimal balance between these factors. For example, as noted by \cite{ashok_arxiv24_ConGenBench}, prompts remain effective in many cases, suggesting that prompt-based methods should not be overlooked. While innovative methods involving latent space manipulation and decoding-phase interventions are promising, the ultimate criterion should be their effectiveness. Researchers should select the most suitable method based on specific application scenarios to achieve the best generation outcomes.

\subsubsection{\textbf{Expanding the Diversity of Testing Tasks}}
Current testing tasks primarily focus on aspects such as toxicity, emotion, topics, and lexicon, with relatively limited evaluations of style and form. Future research should broaden the diversity of testing tasks to include aspects like linguistic style, narrative structure, and pragmatic functions. Introducing these varied testing tasks would allow for a more comprehensive evaluation of the performance and practicality of CTG models.

\subsubsection{\textbf{Maximizing LLM Capabilities When Comparing Baselines}}
When conducting experimental testing, researchers should not limit themselves to traditional CTG methods. With the advancement of LLM technology, it is essential to actively incorporate various existing prompt-based techniques to fully leverage their CTG capabilities. This approach will help in thoroughly evaluating the effectiveness of different methods, ensuring that the chosen baselines are more representative and practical, thereby identifying the optimal solution.

\section{Conclusion}
\label{sec:conclusion}

This paper reviews the latest research advances in the field of Controllable Text Generation (CTG) for Large Language Models (LLMs) and systematically defines the basic concepts, addressing both control conditions and text quality requirements. The paper introduces a new task classification approach, categorizing CTG tasks into content control (or linguistic control/hard control) and attribute control (or semantic control/soft control).

The paper provides a detailed review of various CTG methods. During the training phase, key methods include retraining or fine-tuning pre-trained models and employing reinforcement learning strategies to optimize generation quality and control precision. In the inference phase, commonly used techniques involve guiding generation through prompt engineering, manipulating the latent space for precise control, and intervening during decoding to adjust the output text.

The paper also explores various evaluation methods for CTG and highlights the wide application of CTG technology across multiple vertical domains and general tasks. The challenges faced by the CTG field, including improving quality, optimizing control precision, and enhancing inference efficiency, are discussed, along with future research directions and appeals.

In conclusion, this paper provides a comprehensive review of the core concepts, technical methods, evaluation approaches, and practical applications in the field of controllable text generation, identifying current research challenges and proposing future development directions. It aims to serve as a systematic reference and guide for research exploration in controllable text generation.

\begin{acks}
This work was supported by the National Natural Science Foundation of China (Grants No. 62072463 and 71531012), the National Social Science Foundation of China (Grant No. 18ZDA309), the Research Seed Funds of the School of Interdisciplinary Studies at Renmin University of China, and the Opening Project of the State Key Laboratory of Digital Publishing Technology at Founder Group. 
\end{acks}

\bibliographystyle{ACM-Reference-Format}
\bibliography{bib/retrain,bib/ft,bib/rl,bib/prompt,bib/latent,bib/decoding,bib/evaluation,bib/others}

\appendix

\end{document}